\documentclass[sigconf]{acmart}

\AtBeginDocument{%
  \providecommand\BibTeX{{%
    \normalfont B\kern-0.5em{\scshape i\kern-0.25em b}\kern-0.8em\TeX}}}

\usepackage{multirow}
\usepackage{siunitx}
\usepackage{enumitem}

\usepackage{stfloats}

\copyrightyear{2024}
\acmYear{2024}
\setcopyright{acmlicensed}\acmConference[SIGSPATIAL '24]{The 32nd ACM
International Conference on Advances in Geographic Information
Systems}{October 29-November 1, 2024}{Atlanta, GA, USA}
\acmBooktitle{The 32nd ACM International Conference on Advances in
Geographic Information Systems (SIGSPATIAL '24), October 29-November 1,
2024, Atlanta, GA, USA}
\acmDOI{10.1145/3678717.3691266}
\acmISBN{979-8-4007-1107-7/24/10}

\begin{document}

%%
%% The "title" command has an optional parameter,
%% allowing the author to define a "short title" to be used in page headers.
\title[Learning to Estimate Package Delivery Time in Mixed Imbalanced Delivery and Pickup Logistics Services]{Learning to Estimate Package Delivery Time in Mixed Imbalanced Delivery and Pickup Logistics Services}

%%
%% The "author" command and its associated commands are used to define
%% the authors and their affiliations.
%% Of note is the shared affiliation of the first two authors, and the
%% "authornote" and "authornotemark" commands
%% used to denote shared contribution to the research.
\author{Jinhui Yi}
\orcid{0000-0002-2339-2698}
% \authornotemark[1]
\affiliation{%
  \institution{Department of Electronic Engineering, BNRist, Tsinghua University \& JD Logistics}
  % \institution{JD Logistics}
  \streetaddress{30 Shuangqing Rd}
  %\city{Haidian Qu}
  \state{Beijing}
  \country{China}
  \postcode{100080}
}
\email{yi-jh16@tsinghua.org.cn}
%%%%%%%%%%%%%%%%%%%%%%%%%%%%%%%%%%%%%%%%%%%%%%%%%%%%

\author{Huan Yan$^{\ast}$}
\affiliation{%
  \institution{Department of Electronic Engineering, BNRist, Tsinghua University}
  \streetaddress{30 Shuangqing Rd}
  \city{Beijing}
  \country{China}
%\email{larst@affiliation.org}
}
\email{yanhuanthu@gmail.com}

\author{Haotian Wang}
\affiliation{%
  \institution{JD Logistics}
  \city{Beijing}
  \country{China}
}
\email{wanghaotian18@jd.com}

\author{Jian Yuan}
\affiliation{%
 \institution{Department of Electronic Engineering, BNRist, Tsinghua University}
 \streetaddress{30 Shuangqing Rd}
 %\city{Haidian Qu}
 \state{Beijing}
 \country{China}
 }
 \email{jyuan@tsinghua.edu.cn}

 \author{Yong Li$^{\ast}$}
\affiliation{%
 \institution{Department of Electronic Engineering, BNRist, Tsinghua University}
 \streetaddress{30 Shuangqing Rd}
 %\city{Haidian Qu}
 \state{Beijing}
 \country{China}
 }
 \email{liyong07@tsinghua.edu.cn}

%\thanks{$*$ Corresponding author. Email: yanhuan@tsinghua.edu.cn}
%%
%% By default, the full list of authors will be used in the page
%% headers. Often, this list is too long, and will overlap
%% other information printed in the page headers. This command allows
%% the author to define a more concise list
%% of authors' names for this purpose.
\renewcommand{\shortauthors}{Jinhui Yi, Huan Yan, Haotian Wang, Jian Yuan, \& Yong Li.}
\thanks{$*$ Huan Yan and Yong Li are the corresponding authors.}
%%
%% The abstract is a short summary of the work to be presented in the
%% article.
\begin{abstract}
%写作思路：第一句背景介绍物流快速发展，并引出准确估计送达时间的重要性；接着一句话介绍已有工作的大致方法，并引出有三大挑战它们没能很好处理（每个挑战一句话概况其中核心问题）；然后，介绍所提出的模型，并针对每个挑战介绍其中所设计的方法（要从方法角度介绍其核心思想，不要按照每个模块方式去介绍）；最后介绍实验验证结果和部署情况。
%In recent years, the logistics industry has become increasingly crucial in people's lives for its delivery and pickup services.
Accurately estimating package delivery time is essential to the logistics industry, which enables reasonable work allocation and on-time service guarantee. This becomes even more necessary in mixed logistics scenarios where couriers handle a high volume of delivery and a smaller number of pickup simultaneously.
However, most of the related works %of package delivery time estimation 
treat the pickup and delivery patterns on couriers' decision behavior equally, neglecting that the pickup has a greater impact on couriers' decision-making compared to the delivery due to its tighter time constraints.
In such context, we have three main challenges: 
1) multiple spatiotemporal factors %like distance and remaining time 
are intricately interconnected, significantly affecting couriers' delivery behavior;
2) pickups have stricter time requirements but are limited in number, making it challenging to model their effects on couriers' delivery process;
3) %there is a lack of exploration on mining couriers' spatial mobility patterns, which are also the determinant of couriers' delivery behavior. 
couriers' spatial mobility patterns are critical determinants of their delivery behavior, but have been insufficiently explored.
To deal with these, we propose TransPDT, a Transformer-based multi-task package delivery time prediction model.
We first employ the Transformer encoder architecture to capture the spatio-temporal dependencies of couriers' historical travel routes and pending package sets. 
Then we design the pattern memory to learn the patterns of pickup in the imbalanced dataset via attention mechanism. 
We also set the route prediction as an auxiliary task of delivery time prediction, and
incorporate the prior courier spatial movement regularities 
%by constructing two mobility and spatial matrices 
in prediction. 
Extensive experiments on real industry-scale datasets demonstrate the superiority of our method. A system based on TransPDT is deployed internally in JD Logistics to track more than 2000 couriers handling hundreds of thousands of packages per day in Beijing, 
and the average daily delivery timely rate of deployed stations is 0.68\% higher than the non-deployed stations.

\end{abstract}

% % \vspace{-\baselineskip}
% \vspace{-100cm}
%%
%% The code below is generated by the tool at http://dl.acm.org/ccs.cfm.
%% Please copy and paste the code instead of the example below.
%%


%%
%% Keywords. The author(s) should pick words that accurately describe
%% the work being presented. Separate the keywords with commas.
%% \vspace{-0.5cm}

%% A "teaser" image appears between the author and affiliation
%% information and the body of the document, and typically spans the
%% page.

% \begin{teaserfigure}
%   \includegraphics[width=\textwidth]{sampleteaser}
%   \caption{Seattle Mariners at Spring Training, 2010.}
%   \Description{Enjoying the baseball game from the third-base
%   seats. Ichiro Suzuki preparing to bat.}
%   \label{fig:teaser}
% \end{teaserfigure}

% \received{20 February 2007}
% \received[revised]{12 March 2009}
% \received[accepted]{5 June 2009}

%%
%% This command processes the author and affiliation and title
%% information and builds the first part of the formatted document.
\maketitle
\vspace{-0.3cm}
\section{Introduction}
\label{intro}
%~\cite{yi2023deepsta}
The rapid development of e-commerce has led to the significant increase in logistics demands, especially in mixed imbalanced delivery and pickup services where %couriers handle a high volume of delivery tasks and a smaller number of pickup tasks for customers simultaneously.
couriers handle a significantly higher volume of deliveries than pickups simultaneously.
There are over $200,000$ couriers in JD Logistics, with each of them handling an average of over a hundred deliveries and dozens of pickups per day
~\footnote{https://www.jingdonglogistics.com/}.
Due to the relatively small number and strict time requirements, most pickup tasks should be finished within an hour meeting customers' expectations, thus the logistics industry pays more attention to 
%greater emphasis on the successful completion of deliveries.In the logistics industry, there has been an increasing emphasis on 
estimating package delivery time.
The accurate estimation can alleviate customers' waiting anxiety in face-to-face delivery services, allow the companies to optimize task allocation more effectively and take intervention in couriers' work assignments to ensure on-time delivery.
JD Logistics estimates that a delayed package will result in a loss of over 20 RMB (about 3 USD). 
Moreover, delayed packages will also impact the company's competitive advantage, as customers tend to prefer companies that can provide faster delivery.
%Besides, delayed delivery will also customers will prefer companies that can provide faster delivery, thus choose JD Logistics over other logistics companies to deliver packages in the future.

\begin{figure}[t]
\centering
\includegraphics[width=0.9\linewidth]{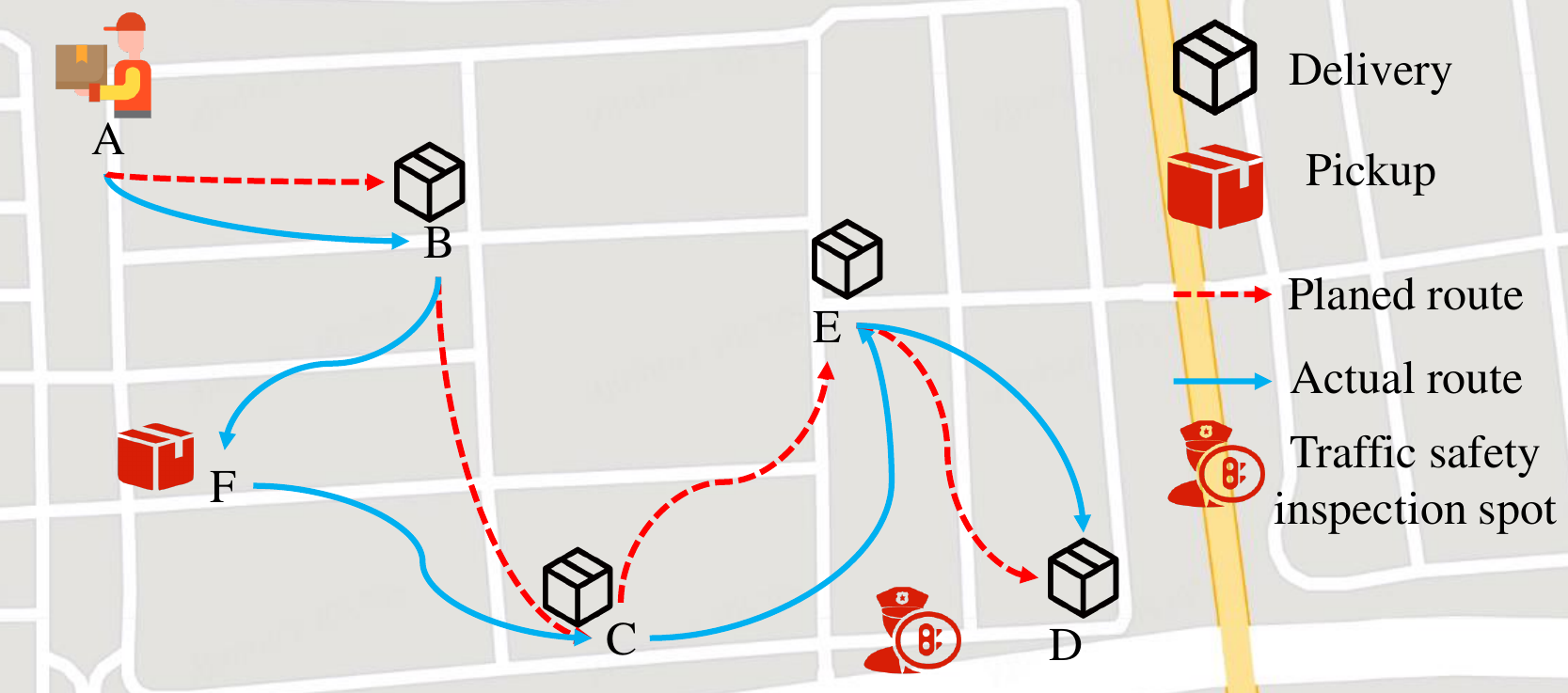}
% \vspace{-4mm}
\caption{The mixed imbalanced delivery and pickup services.
} 
\label{mixed}
%% \vspace{-2mm}
\end{figure}

However, accurately estimating package delivery time in such mixed imbalanced delivery and pickup services is not an easy task.
As shown in Figure~\ref{mixed}, delivery packages $\{A,B,C,D,E\}$ are distributed across multiple locations, and couriers plan their routes based on multiple factors like distance, remaining time and personal habits. Additionally, the uncertain pickup task ${F}$ has stricter time requirements, making couriers pause delivery and prioritize pickup frequently. 
As summarized in Table~\ref{dif}, existing research on package delivery time prediction can be categorized into two main categories.
% \vspace{-1mm}
% \vspace{-2mm}

\vspace{-0.3cm}
\begin{table}[ht!]
\centering
\caption{Differences among online e-commerce scenarios.}\label{dif}
\vspace{-0.3cm}
\resizebox{\columnwidth}{!}{%
\begin{tabular}{|c|c|c|c|c|}
\hline
\multirow{2}{*}{} & \multicolumn{2}{c|}{\textbf{Single service}} & \multicolumn{2}{c|}{\textbf{Mixed service}} \\
\cline{2-5}
 &  \textbf{Delivery} & \textbf{Pickup}  &  \textbf{Pair-wise delivery \& pickup} &  \textbf{Imbalanced delivery \& pickup} \\
\hline
% \multirow{2}{*}{\textbf{Application}} & KFC & Cainiao Ltd& Online food ordering& JD Logistics, SF Express \\
% \cline{5-5}
%  & & & & \\
\textbf{Application} & KFC & Cainiao Ltd& Online food ordering& JD Logistics, SF Express \\
\hline
% \textbf{Task balance} &balanced &  balanced &balanced& imbalanced \\
% \hline
\multirow{2}{*}{\textbf{Daily number}} &\multirow{2}{*}{\centering high} &  \multirow{2}{*}{\centering high} & {\centering high (delivery)} & {\centering high (delivery)}\\
 & &  & {\textbf{high (pickup)}}& {\textbf{low (pickup)}}\\
\hline
\multirow{2}{*}{\textbf{Uncertainty}} &\multirow{2}{*}{\centering low} &  \multirow{2}{*}{\centering high} & {\centering high (delivery)} & {\centering low (delivery)}\\
 & &  & {{high (pickup)}}& {{high (pickup)}}\\
% \textbf{Uncertainty} & low & high& high(delivery) & low(delivery) \\
%  &  & & high(pickup) & high(pickup) \\
\hline
\end{tabular}%
}
%% \vspace{-2mm}
\end{table}
\vspace{-3mm}

\begin{figure}[hbp]
\centering
\includegraphics[width=0.95\linewidth]{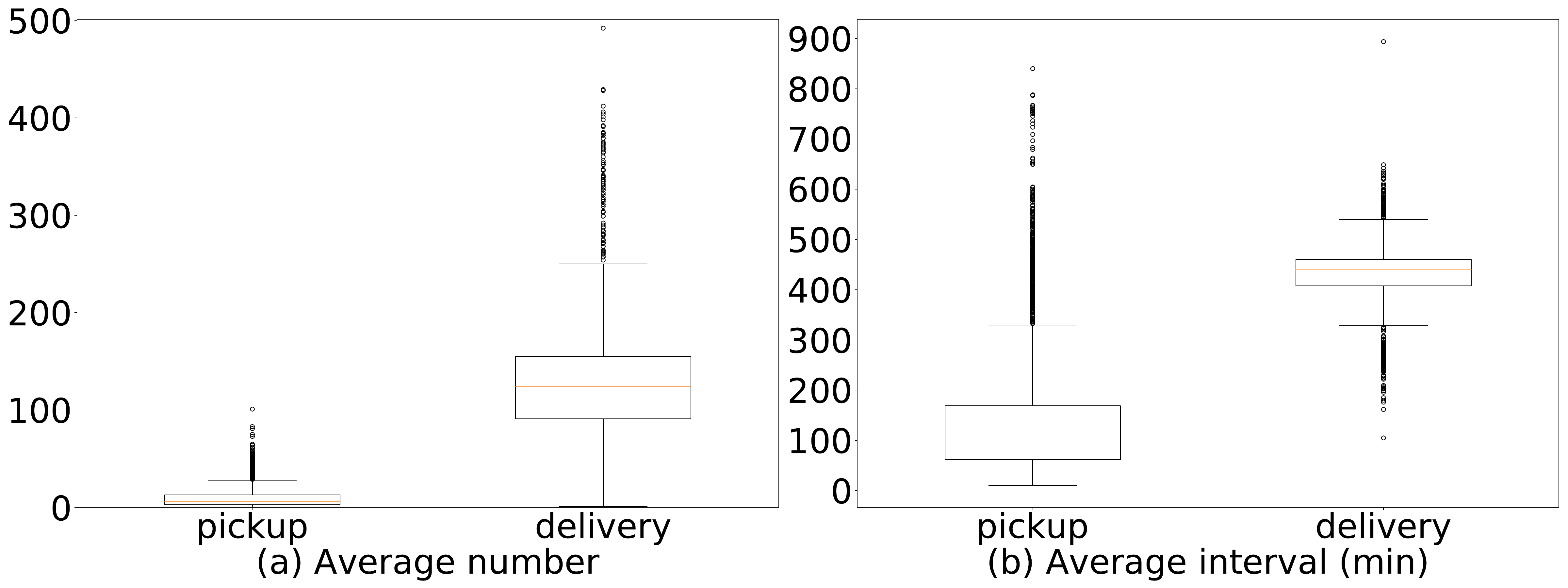}
\vspace{-2mm}
\caption{The statistical results of JD Logistics from December 2022 to July 2023, in Beijing of (a) the average daily number of delivery and pickup completed by each courier, (b) the average time interval between packages' dispatched time and promised finish time.
} 
% \vspace{-19mm}
\label{number_remaing}
\end{figure}

The first category is the single service mode, where
%In the majority of existing research on package completion time prediction, 
couriers are assumed typically to offer a single service of delivery~\cite{wu2019deepeta, ruan2022service} or pickup~\cite{wen2021package,wen2022deeproute+,wen2023enough,wen2023modeling} exclusively.
The research on delivery often focuses on uncovering relatively fixed patterns from couriers' historical delivery behaviors, 
%but overlooks the influence of uncertain pickups on the delivery process. Conversely, 
while studies on pickup focus on mining correlations between packages and locations.
% Both of them lack the capability to model the inter-influence between the pickup and delivery tasks.  
The second category is the mixed service mode where couriers offer both delivery and pickup services, and can be further subdivided into two types, \emph{i.e.}, the pair-wise pickup \& delivery scenario~\cite{feng2023ilroute,gao2021deep,zhu2020order} and the imbalanced pickup \& delivery scenario.
%In practice, it is quite common for express companies like JD Logistics and SF Express to provide mixed delivery and pickup services.
%There are inherent distinctions between the two types, and the main difference lies in the number of pickup samples. 
The typical scenario for the first type is the online food ordering. 
The pickup and delivery activities of each package always occur in pairs, resulting in abundant pickup samples for analysis. 
On the contrary, for the second type, the pickup and delivery exhibit a significant imbalance in terms of quantity.
%which makes the conventional food ordering methods generalize the majority delivery samples easily, while failing to capture patterns of minority pickup samples.
% Besides, the two scenarios also differ in the spatio-temporal constraints.
%The food takeaway service has the pick-up-then-deliver constraint, the picked-up packages will turn into drivers' pending delivery tasks, and continuously influence drivers' future decisions. The logistics couriers, on the other hand, receive all the deliveries in the logistics station before starting work, and they take pickups back to stations without the need for further delivery. 
%As a result, the methods for online food ordering exhibit poor performance due to their failure to consider the impact of limited pickup tasks.
Most aforementioned methods ~\cite{feng2023ilroute,gao2021deep} treat the impacts of pickup and delivery on couriers' decision making equally, which can not be applied to this service type.
Formally, there are three key challenges:

%However, accurately estimating package delivery time in mixed delivery and pickup service is not an easy task, where couriers handle multiple delivery tasks and several pickup tasks every day.
\begin{itemize}
\item \textbf{Complex spatio-temporal correlations in courier decision making.}
Since couriers prioritize packages with shorter remaining time or closer distance proximity, the spatio-temporal constraints of packages such as location, the promised time and remaining time, 
%which largely determine package delivery routes, thereby directly affecting the delivery time. 
will largely determine package delivery route and time.
%In terms of time, couriers prioritize packages with shorter remaining time. In terms of space, they finish packages in order of distance proximity. 
Nonetheless, it is challenging to capture such complicated spatio-temporal relations effectively.
%many existing approaches fail to model such complicated spatio-temporal relationships effectively. 
%DeepETA~\cite{wu2019deepeta} directly predicts the package delivery time without considering their delivery order. FDNet~\cite{gao2021deep} neglects the spatio-temporal dependencies among packages.
%Thus, it is challenging to capture such spatio-temporal relations.

%是不是改个名，得突出样本数少？
\item \textbf{Limited uncertain pickup impact on couriers' delivery decision making.}
%这条是否给个图，表明收到揽件需求后快递员偏离原路线前往揽件？
As presented in Figure~\ref{mixed}, delivery tasks are dispatched before courier starts work. He/She planned to deliver $C$ after $B$, 
%the route as $A \rightarrow B \rightarrow C \rightarrow E \rightarrow D$. 
but switches for $F$ first after receiving the new pickup request.
As shown in Figure~\ref{number_remaing}, 
%shows the difference in number and time requirement between delivery and pickup. 
couriers handle relatively few pickups compared to the deliveries every day, and the average remaining processing time of pickup is quite less than delivery. %less than two hours, while that of the delivery is more than seven hours. 
%Such stricter time requirements necessitate couriers to respond promptly upon receiving customer pickup requests. 
%Such uncertain pickup requests call for frequent route re-planning. %of courier's pre-determined routes.
%When studying how pickups affect couriers' delivery behaviors 
 However, the related methods~\cite{cai2023m,wen2023modeling} have limitations in effectively capturing the patterns of minority pickups of imbalanced samples. 

\item \textbf{Insufficient modeling of couriers' spatial mobility preferences.}
%这条是否给个图？
The spatial mobility preferences refer to the regularities of couriers' transitions between locations~\cite{wen2022deeproute+,zhang2023efficient}. 
An example is shown in Figure~\ref{mixed}, 
%after completing task $C$, 
although the pending tasks $D$ and $E$ share similar spatio-temporal constraints, most couriers at $C$ 
%in the real world 
will choose to deliver $E$ first then $D$ %the route $C \rightarrow E \rightarrow D$ rather than $C \rightarrow D \rightarrow E$ 
due to the traffic safety inspection spot between $C$ and $D$ which will take more time. 
%The logistics industry has massive amounts of data on human mobility, which can be integrated into the model for performance improvement.
However, the importance of such information has been largely overlooked by existing studies ~\cite{wen2023enough,wen2021package}.
\end{itemize}

\vspace{-4mm}
\vspace{-5mm}

To tackle the above challenges, we propose a Transformer-based multi-task learning package delivery time prediction model, named TransPDT, to estimate the package delivery time.
%prediction as the major task, and the completion route prediction as the auxiliary task for better generalization, 
%which consists of three modules.
%The first package correlation extraction module is designed to address the first challenge, which 
First, the model utilizes the Transformer architecture to capture the spatio-temporal dependencies in couriers' historical travel route and pending package set, respectively.
Next, we design the pattern memory to learn and store the regularities in the pickup's impact on delivery from limited samples via the attention mechanism.
In particular, we set the route prediction as the auxiliary task, together with prior knowledge of couriers' spatial mobility patterns, to enhance time prediction performance. 
%To be specific, we first generate the intermediate probabilities of packages at each step with LSTM and attention layer. Then by constructing two mobility and spatial location-pair matrices from couriers' historical records, we predict the route and time of pending packages subsequently.

The main contributions of this article are summarized as follows:
\begin{itemize}
\item We propose a Transformer-based multi-task framework for the package delivery time estimation in the mixed delivery and pickup services, which incorporates route prediction as an auxiliary task to enhance the performance of time prediction.

\item To capture couriers' spatial movement preferences, we integrate couriers' prior spatial and mobility knowledge into the package route and time prediction.

\item To learn the pickup influencing patterns on delivery with imbalanced training samples, we design the pattern memory to learn and store patterns of the minority pickup tasks, and employ the attention mechanism to model such impact.

\item Extensive experiments on real-world datasets 
show our model outperforms its best competitors by 3.54\% in RMSE and 18.04\% in MAPE.
The model has been deployed internally in JD Logistics to serve the logistics station managers with work allocation and on-time service guarantee.
\end{itemize}
% 本文贡献：（1）提出基于Transformer的包裹送达时间预测模型，该模型同时对快递员历史配送行为和未来配送任务进行建模。并且采用多任务形式，将预测快递员配送顺序作为辅助任务，提升预测的准确性。
% （2）提出XX方法建模快递员配送决策受多重空间因素影响
% （3）提出XX方法建模实时的揽收对快递员配送决策的复杂影响
% （4）实验验证模型的有效性。并且在JD平台部署，服务XX城市，XX个快递员。
\vspace{-4mm}
\section{RELATED WORKS}
\label{Sec:related_work}
% % \vspace{-0.5mm}
\subsection{Package Delivery Time Prediction in Logistics}
\label{sec:Logistics}
%下面这段话是不是放到intro里边去？？
The logistics industry usually offers two modes of service: delivery and pickup. 
For the delivery service~\cite{wu2019deepeta,de2021end,zhou2023inductive,zhang2023dual}, the sequence of package completion follows a relatively fixed pattern. 
Nevertheless, for the pickup service~\cite{ruan2022service,wen2023enough}, it is quite challenging to discover sequential patterns, due to the unpredictability of time and the randomness of locations of pickup requests. 
Most of the previous solutions focus on predicting package arrival time under a single service mode. 
Wu et al.~\cite{wu2019deepeta} proposed DeepETA, which employs the Long Short-Term Memory (LSTM) to model the sequential features of delivery routes, and searches for similar historical routes and sets of pending packages in the historical data.
%to predict package delivery time. 
However, their method fails to consider the impact of uncertain pickup on the delivery. Furthermore, they neglected the delivery order, which is a vital factor for the delivery time.
Wen et al.~\cite{wen2023enough} introduced RankETPA, which first employs an attention-based recurrent neural network to predict the pickup sequence, and then predicts the pickup time via Transformer encoder architecture to mine the correlations among the pending pickups. However, they overlooked couriers' delivery mobility regularities, and failed to model the complicated relationships between pickups and deliveries.

%In the real world, it is quite common for logistics companies to require their couriers to offer both delivery and pickup services simultaneously. 
In the mixed delivery and pickup service scenario, the occurrence of real-time pickups can have a substantial impact on delivery.
%Nevertheless, in the context of the mixed services scenario, 
Besides, when determining the delivery sequence, couriers not only prioritize time-sensitive pickups but also consider their own preferences.
%, which have not been modeled in their approach.
% \vspace{-4mm}
\subsection{Estimated Time of Arrival in Traffic}
\label{sec:Traffic}
%关键得说明和我们任务的区别: ranketa+last mile，总共三点
With the rapid development of online ride-hailing and navigation, the ETA (Estimated Time of Arrival) problem has become popular which refers to the travel time estimation between the origin and destination.
Existing research on ETA can be grouped into two main categories. % based on their dependence on route information.
%这里是否加引用？
The first category is the OD-based method~\cite{han2023ieta,zhang2020real,hu2020stochastic,yuan2020effective,wang2019simple,yan2022jointly,jin2021hierarchical}, where the input only consists of an origin-destination without explicit route information.
For instance, ~\cite{han2023ieta} proposed iETA, which achieves better estimation performances by combining 
real-time traffic data and historical traffic knowledge under the incremental learning setting.
% ~\cite{zhang2020real} designed a multi-layer graph convolutional networks-based model, which extracts traffic flow information from road surveillance cameras for travel time estimation.
%这个人也使用了aux任务，我们之后可以写下
%~\cite{yuan2020effective} designed an auxiliary task to bind each OD pair to corresponding historical trajectories in the training stage to boost the performance of the model. 
The second category is the route-based method~\cite{fu2020compacteta,chen2022interpreting,fang2020constgat,jin2022stgnn,wang2018learning,sui2024congestion}, which consists of a two-stage process: first predicting the optimal route and then estimating the arrival time. By incorporating road information into the estimation, such an approach enhances time prediction performance. 
For example, ~\cite{chen2022interpreting} proposed HierETA, a hierarchical self-attention network which learns the comprehensive representation of trajectory to estimate time of arrival. 
%~\cite{fu2020compacteta} applies a graph attention network to capture spatial-temporal dependencies, and introduces positional encoding to process the sequential travel route. 
% ConSTGAT~\cite{fang2020constgat} adopts convolutions over local windows to better capture a route's contextual information.

The delivery time prediction differs from the typical ETA problem significantly, making it quite challenging to directly borrow from existing ideas.
Firstly, the typical ETA prediction takes one origin with a single destination as input, whereas our input is an origin with multiple destinations (packages) ~\cite{wen2023enough}. Additionally, due to the influence of uncertain events like pickup, rest and meal, the order of couriers visiting multiple destinations still remains unknown when predicting.
% % \vspace{-4mm}
\subsection{Route Prediction in Online Business}
Logistics and food takeaway are two main representative online businesses. In recent years, the prediction of the arrival routes of packages or food orders has gained increasing attention, due to its potential in optimizing employee task allocation~\cite{wen2021package,wen2022deeproute+,gao2021deep,feng2023ilroute,wen2023survey,wen2022graph2route,zhang2023delivery}. 
Moreover, the outcome of route prediction is fundamental for the time prediction task.
DeepRoute+~\cite{wen2022deeproute+} extracts spatio-temporal constraints of unpicked packages via Transformer architecture, and predicts the pickup route with attention-based recurrent neural network (RNN). 
ILRoute~\cite{feng2023ilroute} utilizes the Graph-based imitation learning method for route prediction in online food ordering.
%FDNET~\cite{gao2021deep} leverages an RNN and attention-based route prediction module to enhance the performance of the time prediction module, while it neglects the correlations between packages.
CP-Route~\cite{wen2023modeling} integrates the human mobility knowledge into pickup route prediction though, it fails to model the impact of pickup on delivery, and does not capture the sequential pattern of couriers' historical travel route.
M\textsuperscript{2}G4RTP~\cite{cai2023m} solves the route and time prediction in instant logistics, by applying the graph attention network to model spatial correlations between locations. However, it overlooks the movement regularities from the perspective of couriers. 
DRL4Route~\cite{mao2023drl4route} introduces reinforcement learning into route prediction under the pick-up scenario primarily, 
without modeling the impact of uncertain events on the future travel route.
In conclusion, in the above works, couriers are assumed to work in single service mode, which does not match our mixed service mode and fails to consider the interaction between different services. 
%And most of them neglect to model the courier's mobility preferences.

\section{PRELIMINARIES}
%这里主要是三种定义吧：包裹--路线&时间--问题定义
\textbf{Definition 1 (Package).}
In the logistics industry, packages can be categorized into two types: delivery and pickup. The pickup task is typically required to be completed within a shorter period, while the delivery task has a more flexible deadline.
%这里是否画一张图，表明揽收的时限要求比派件短？
Each package $o$ has several properties, \emph{i.e.}, $o=(type, loc, tr, feat)$, where $type$ denotes the type (delivery or pickup) of the package, $loc$ is the location (latitude and longitude) of the package, $tr$ is the time requirements (required completion time, actual completion time, courier's dispatched time) of the task, and $feat$ represents other information (weight, volume, etc.) of the order.

\noindent\textbf{Definition 2 (Historical package route).}
The historical package route $H_t=\{o_1,..., o_m\}$, refers to the sequence of packages that the courier already completed before time $t$, where $m=m_d+m_p$ is the number of finished delivery and pickup tasks and $o_i$ represents the $i$-th finished package.

\noindent\textbf{Definition 3 (Future package route).}
At time $t$, given the courier's pending package set $F_t=\{o_1,..., o_n\}$ where $n=n_d+n_p$ denotes the number of pending delivery and pickup packages, his/her future package route $\pi$ is the permutation of $F_t$:
\begin{equation}
    \pi(F_t)=(\pi_1,...,\pi_n),
\end{equation}
where $\pi_i \in \{1,...,n\}$ and $\pi_i \neq \pi_j$ if $i \neq j$. For instance, $\pi_i = j$ means the $i$-th package in the completion route is the $j$-th package in $F_t$.

\noindent\textbf{Problem Definition.}
Given the courier's historical package route $H_t$, and pending \emph{delivery} and \emph{pickup} package set $F_t$ at time $t$,
the objective is to predict the arrival time of \emph{delivery} packages in $F_t$:
\begin{equation}
    \phi(H_t,F_t) \rightarrow y=\{y_1,...,y_{n_d}\},
\end{equation}
where $y_i$ is the time interval between $t$ and the actual delivered time of the $i$-th delivery package in $F_t$, $n_d$ is the number of delivery packages, and $\phi$ denotes the mapping function to learn.
%\textbf{Lack of sufficient training data.}
%The courier handles multiple delivery and pickup tasks $O=\{o_1^d,o_2^d,o_3^p,...\}$ every day, where the superscript $d$ denotes delivery, and $p$ denotes pickup. Given the information of finished packages $O_{history}=\{o_1^d,o_2^d,o_3^p,...,o_{n1}^d\}$ and to be finished packages $O_{future}=\{o_1^d,o_2^p,o_3^d,...,o_{n2}^d\}$ of courier $c$, our goal is to predict the delivery time $T_{d}=\{t_1,t_3,...,t_{n2}\}$ for packages to be delivered in $O_{future}$.
% 这里还是语焉不详，得说明揽收不用预测。
% 先不加入加入stay point/AOI
% 干脆改成p_1=(geo,time,other feature)的表达形式

%在这里介绍dataset截断
\section{Method}
\subsection{Model Overview}
\begin{figure}[htbp]
\centering
\includegraphics[width=\linewidth]{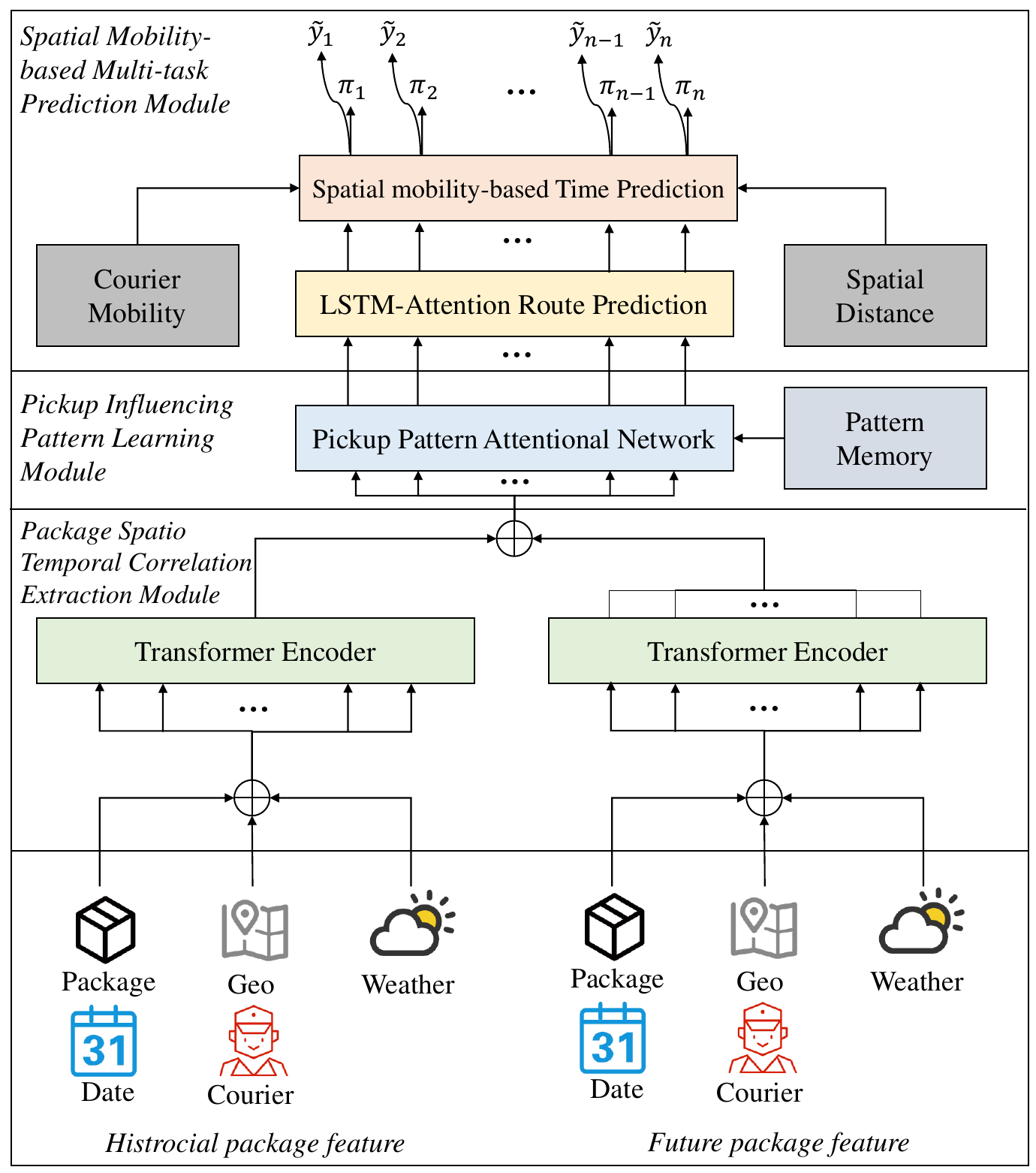}
%% \vspace{-7mm}
\caption{Model architecture overview.
} 
\label{overrall}
%% \vspace{-6mm}
\end{figure}

An overview of TransPDT is shown in Figure~\ref{overrall}, which is made up of three modules: the package spatio-temporal correlation extraction module, the pickup influencing pattern learning module, and the mobility-based package prediction module. 
%The package correlation extraction module employs the Transformer Encoder to model the features of finished tasks, and capture the correlations between packages to be delivered.
The package spatio-temporal correlation extraction module is designed to tackle the first challenge, which is divided into two branches, with each branch employing a Transformer encoder to model the historical package route and the pending package set, respectively.
The pickup influencing pattern learning module is designed to overcome the second challenge, which learns the pattern of pickup tasks with imbalanced datasets via the attention mechanism.
The spatial mobility-based multi-task prediction module is designed to solve the third challenge. It first utilizes the attention-based LSTM to predict the intermediate probability of each package the courier will deliver or pick up in each step.
%, and generate the courier's future route.
%The human mobility learning module enhances the model's performance by extracting the spatial relationships between areas of interest (AOIs) and learning human mobility patterns based on historical logistics data.
It then integrates the prior knowledge of spatial relation and movement regularity between locations, to subsequently generate the courier's future package route as the auxiliary task and corresponding delivery time as the main task.
%% \vspace{-3mm}

\subsection{Package Spatio-Temporal Correlation Extraction Module}
\begin{figure}[tbp]
\centering
\includegraphics[width=\linewidth]{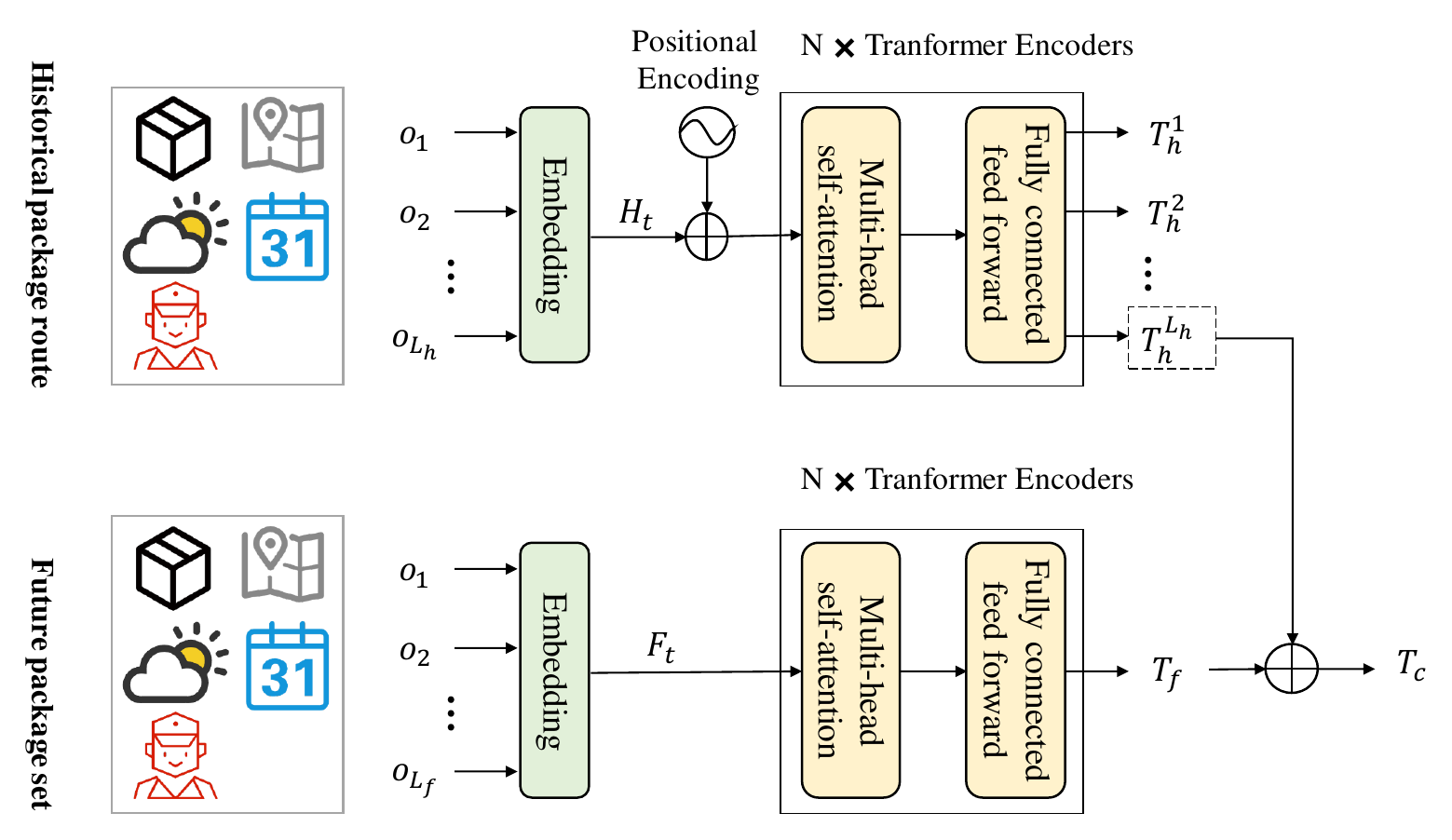}
% % \vspace{-5mm}
\caption{Package spatio-temporal correlation extraction module.
} 
\label{Transformer}
% \vspace{-6mm}
\end{figure}

This module generates representation for historical packages and pending packages respectively as depicted in Figure~\ref{Transformer}.
%下面这句话得改下，而且不一定要引用
%As mentioned in Section ~\ref{Sec:}, the distribution of AOIs visited previously and the arrangement of pending packages actively influence the future itinerary of the courier.
%AOIs the courier visited previously will not be revisited within a short time, and the courier prioritizes the pending packages considering factors like distance, remaining time, and packages' distribution across AOIs. 
The authors in ~\cite{wu2019deepeta} consider human mobility as a probabilistic chain, and future transition patterns can be inferred from historical behaviors.
Besides, the spatio-temporal correlations among pending packages significantly influence the delivery route and time.
Therefore, we divide the module into two branches to model the historical and future package information respectively via the Transformer encoder. 

\textbf{Transformer Encoder.} 
Previous work ~\cite{wu2019deepeta,wen2022deeproute+,wen2023enough} employ RNN and its variants
%the Bidirectional Long short-term memory (BiLSTM) 
to capture the correlations between packages. Couriers usually handle a large number of packages, and RNN-based methods have difficulty in capturing long-distance dependencies and is limited in parallelization.
On the contrary, Transformer~\cite{vaswani2017attention} learns the relation between any input package pair with the self-attention mechanism, and its output is invariant to the input order, which makes it more suitable for handling multiple pending packages with unknown sequences.
Each Transformer encoder block is composed of two sub-layers~\cite{vaswani2017attention}. The first is a multi-head self-attention layer, and the second is a %position-wise 
fully connected feed-forward layer.
The first layer adopts the self-attention mechanism to model relations of input tokens, which can be formulated as:
\begin{equation}
\text{Attention}(\mathbf{Q, K, V}) = \text{Softmax}\left(\frac{\mathbf{QK^T}}{\sqrt{d_k}}\right)\mathbf{V},
\end{equation}
where $\mathbf{Q, K, V}$ are the queries, keys and values,% projected from the $M_h$ or $M_f$, 
and $d_k$ is the dimension of $K$. Then the multi-head attention allows the model to learn information from different representation subspaces.
% , which can be formulated as:
% \begin{equation}
% \text{MultiHead}(\mathbf{Q, K, V}) = \text{Concat}(\text{head}_1, \text{head}_2, \ldots, \text{head}_h)W^O,
% where \text{head}_i=\text{Attention}(\mathbf{Q}W_i^Q, \mathbf{K}W_i^K, \mathbf{V}W_i^V),
% \end{equation}
The outputs of the first layer will then be fed into the second layer consisting of two linear transformations.
Then, we stack multiple blocks together to increase model's depth and non-linearity. 

Besides, the positional encoding enables the model to use packages' sequential information, which can be formulated as:
\begin{equation}
\begin{aligned}
\text{PE}(pos, 2i) &= \sin\left(\frac{pos}{10000^{2i/d_{model}}}\right), \\
\text{PE}(pos, 2i+1) &= \cos\left(\frac{pos}{10000^{2i/d_{model}}}\right),
\end{aligned}
\end{equation}
where $pos$ represents the position of the package, $i$ is the dimension of the feature of the package and $d_{model}$ denotes model's hidden dimension. 

% % \vspace{-3mm}

\textbf{Two branches.} 
We leverage two Transformer encoders as two branches to capture the spatio-temporal dependencies of historical package route and future package set.
To be specific, we first embed packages' features, including package feature, geographical feature, weather feature, date feature and courier feature, by normalizing numerical features and encoding categorical features with one-hot.
Then the feature matrix of historical package route $\mathbf{H_t} \in \mathbb{R}^{L_h \times D_m}$ and future package set $\mathbf{F_t} \in \mathbb{R}^{L_f \times D_m}$ will be fed into each branch respectively, and the corresponding outputs are $\mathbf{T_h} = (T_{h}^1,...,T_{h}^{L_{h}}) \in \mathbb{R}^{L_h \times d_{model}}$ and the second encoder outputs $\mathbf{T_f} = (T_{f}^1,...,T_{f}^{L_{f}}) \in \mathbb{R}^{L_f \times d_{model}}$, 
where $L_h$ and $L_f$ denote the length of completed and pending package routes, and $D_m$ denotes the dimension of packages' feature.
%, which leverages the self-attention mechanism to capture the dependencies among packages and weigh their importance.
Particularly, the positional encoding is only applied to the historical branch of $H_t$, since the order of pending packages $F_t$ remains unknown in the inference stage. 

% We concatenate the last time step of the output from the first model, which represents the current position of the courier, with the output from the second model.
Then we fuse $T_h$  and $T_f$ by concatenation operation, denoted as $\mathbf{T_c} \in \mathbb{R}^{L_f \times 2d_{model}}$. Note that for the historical representation $T_h$, we only use the representation of the last package, \emph{i.e.}, $\mathbf{T_{h}^{L_h}} \in \mathbb{R}^{d_{model}}$, which integrates courier's overall historical behaviors, and the concatenation is implemented through broadcasting.
%and $d_model$
%是否在这里说明输入特征有哪些？需要说明aoi id已被emb
%改写成two stage
% % \vspace{-3mm}
\subsection{Pickup Influencing Pattern Learning Module}
\begin{figure}[htbp]
\centering
\includegraphics[width=\linewidth]{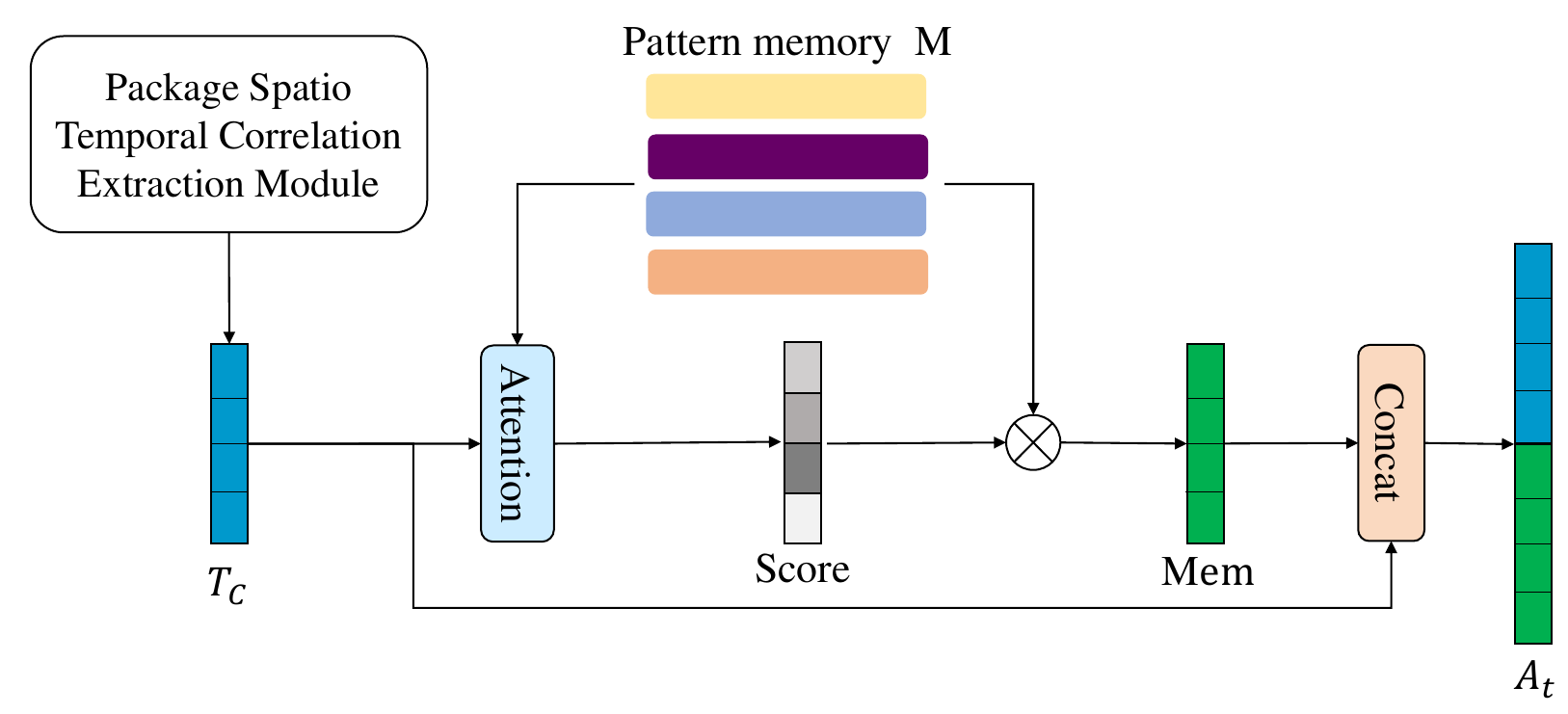}
% % \vspace{-7mm}
\caption{Pickup influencing pattern learning module.
} 
\label{Memory}
% % \vspace{-6mm}
\end{figure}

This module is designed to learn the 
%complex impact of pickup on couriers' delivery 
patterns of pickup tasks 
with limited training samples, as illustrated in Figure~\ref{Memory}. As mentioned in Section ~\ref{intro}, couriers may prioritize pickup tasks due to their tighter time requirements.
Additionally, customers'  uncertain pickup requests lead to frequent route adjustments by couriers.
%the priority of pickup is higher than delivery, thus the courier often changes the planned delivery route after a pickup task is dispatched to him. 
%However, the number of pickup tasks is quite limited compared to the hundreds of daily delivery tasks. The lack of sufficient samples makes it hard to study the impact of pickup on delivery. 
However, the relatively limited number of pickup tasks leads to insufficient training samples, and hinders the performance of traditional data-driven methods.
%To learn patterns from small samples when the sample size is limited, 
Many existing works~\cite{wang2022event,li2023traffic,yi2023deepsta,tang2020joint} learn the patterns of small samples via the attention mechanism. 
Hence, we design a pickup pattern attentional network %via the attention mechanism 
to tackle this challenge.

\textbf{Pickup Pattern Attentional Network.} The network adopts the pattern memory which serves as learnable key-value pairs, to learn and store the patterns of limited pickup packages.
%the impact of pickup on delivery patterns during the training phase.
The weight of pattern memory $\mathbf{M} \in \mathbb{R}^{L_m \times D_m}$ is randomly initialized, where $L_m$ is the number of stored patterns and $D_m$ is the dimension of patterns. 
We use the output of the last module as the query of the pattern memory, to calculate the similarity score between pattern memory $\mathbf{M}$ and pending packages $\mathbf{T_c}$ via the attention mechanism:
\begin{equation}
\mathbf{Score}=Softmax(\mathbf{M} \odot \mathbf{T_c}),
\end{equation}
where $\odot$ denotes the inner product. 
Thus we obtain the representation vector by matrix multiplication:
\begin{equation}
\mathbf{Mem}=\mathbf{Score} \times \mathbf{M},
\end{equation}
Then $\mathbf{Mem} \in \mathbb{R}^{L_f \times 2d_{model}}$ will be concatenated with $\mathbf{T_c}$ together as the output $\mathbf{A_t}=(\mathbf{A_t}^1,...,\mathbf{A_t}^{L_f}) \in \mathbb{R}^{L_f \times 4d_{model}}$ of the module:
\begin{equation}
    \mathbf{A_t}=\mathbf{T_c}||\mathbf{Mem},
\end{equation}
where $\mathbf{A_t}^i$ denotes the representation of the pending package $o_i$, and $||$ denotes concatenation.

% % \vspace{-4mm}
% \vspace{-4mm}

\subsection{Spatial Mobility-based Multi-task Prediction Module}
\begin{figure}[tbp]
\centering
\includegraphics[width=\linewidth]{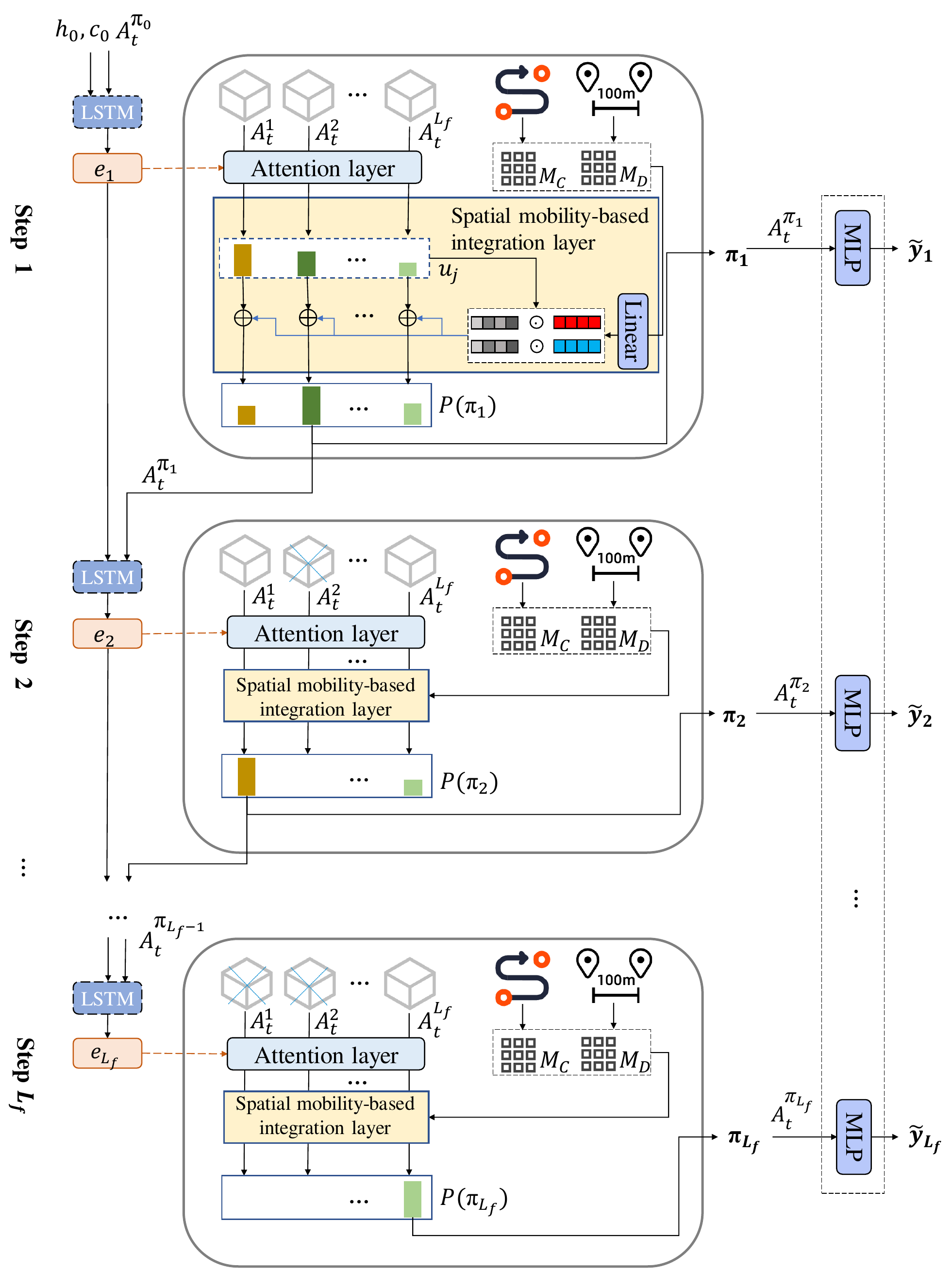}
% % \vspace{-5mm}
\caption{Spatial mobility-based multi-task prediction module.
} 
\label{RPTP}
% % \vspace{-6mm}
\end{figure}

%下面这里是不是把路线formulate一下？
This module is comprised of two components to integrate couriers' spatial mobility preferences into the route and time prediction consecutively as presented in  Figure~\ref{RPTP}.
The first LSTM-Attention route prediction component predicts the intermediate probability of pending packages at each step. The second spatial mobility-based time prediction component learns couriers' spatial mobility regularities and generates packages' final route and time prediction results.
\subsubsection{LSTM-Attention Route Prediction Component}
%generate courier's future delivery and pickup route by calculating 
%This component is designed to calculate the intermediate probability of courier handling each package at each step, based on the output representation of the previous module, which can be formulated as:
%下面要说明只是预测中间变量
%This component forecasts the courier's future package route $\mathbf{\pi}$ by calculating the intermediate probability of the courier handling each package steply, which can be formulated as:
This component 
%forecasts the courier's future package route $\mathbf{\pi}$ by 
calculates the intermediate probability of the courier handling each package step by step, 
which can be formulated as:
\begin{equation}
p(\pi_{j}|\pi_{0},...,\pi_{j-1},\mathbf{A_t}),
\end{equation}
where $j$ denotes the time step of packages.
Inspired by the pointer network~\cite{vinyals2015pointer} which has been widely used in route prediction, we first adopt an LSTM layer to process the sequential package information, and then design an attention layer to forecast the probabilities of packages on each timestamp.

\textbf{The LSTM Layer.} 
%The formulation of the LSTM layer is as follows:
At step $j$, LSTM takes $\mathbf{A_t^{\pi_{j-1}}}$ as input, which denotes the representation of the previous package $\pi_{j-1}$ produced by the last module, as formulated: 
\begin{equation}
\mathbf{e}_j,(\mathbf{h}_j,\mathbf{c}_j)=LSTM(\mathbf{A_t}^{\pi_{j-1}},(\mathbf{h}_{j-1},\mathbf{c}_{j-1})),
\end{equation}
where 
%$v_{i-1} \in \mathbb{R}^{4d_{model}}$is the corresponding feature extracted from $A_t$ of the predicted package $p_{i-1}$ on last timestamp $i-1$, 
$\mathbf{e}_j$, $\mathbf{h}_j$ and $\mathbf{c}_j$ are the output vector, hidden state and cell state of LSTM respectively. 
In particular, at step $j=0$, $\mathbf{h}_0,\mathbf{c}_0$ are randomly initialized and $\mathbf{A_t}^{\pi_{0}}=mean(\mathbf{A_t})$ to represent the whole pending package set.

\textbf{The Attention Layer.} The attention layer calculates the probability for all the pending packages at each time step based on the pointer network. 
We first compute the attention score for all packages on each time step:
\begin{equation}
s_j^i = \begin{cases}
    v^T(\mathbf{W_1}\mathbf{e}_j+\mathbf{W_2}\mathbf{A_t}^i) & \text{if } i \neq \pi_{j'}, \forall j' < j \\
    -\infty & \text{otherwise}
\end{cases}
,
\end{equation}
where $s_j^i$ denotes the attention score of each pending package $o_i$ on step $j$, and $\mathbf{W_1}, \mathbf{W_2}$ and $v$ are learnable parameters, and packages already output in $\mathbf{\pi}$ before $j$ will be masked.
Then we calculate the intermediate probability of packages $\mathbf{u}_j=(u_j^1,u_j^2,...,u_j^{L_f})$ at $j$ via softmax:
\begin{equation}
u_j^i = p(\pi_{j}=i|\pi_{0},...,\pi_{j-1},\mathbf{A_t}) = \frac{e^{s_j^i}}{\sum_{k=1}^{L_f} e^{s_j^k}}.
\end{equation}
%where $u_j=(u_j^1,u_j^2,...,u_j^{L_f})$ denotes the probability of each package at step $j$.

% % \vspace{-3mm}

\subsubsection{Spatial Mobility-based Time Prediction Component}
\label{Mobility component}
This component integrates the prior knowledge of courier mobility preferences and spatial correlations to generate the final prediction of the future package route $\mathbf{\pi}$ and time $y$ successively.

%As discussed in Section ~\ref{intro}, the vast amount of mobility data in the logistics industry can assist us to better model couriers' mobility behavior.Drawing inspiration from ~\cite{feng2020learning} which utilizes the prior knowledge of human mobility via the location relation matrices, 
%Many recent works~\cite{wen2023modeling,feng2020learning,feng2018deepmove} model the spatial movement patterns of individuals between locations via the human mobility data, thereby boosting model's performance.
Previous works~\cite{wen2023modeling,feng2020learning,feng2018deepmove} indicate that there are spatial mobility patterns in the transitions between locations of couriers, which significantly impact packages' delivery order and time. 
~\cite{wen2023modeling} ranks the sequence of packages and locations respectively based on couriers' historical movement records. However, it overlooks the fact that transitions between locations are also influenced by spatial distances and the time slot of the day.
First, we divide the day into 12 time slots, each spanning 2 hours. For each time slot t, we then construct two matrices.
Therefore, we first divide a day into 12 time slots at an interval of 2 hours. Then we construct two original location-pair matrices for $N$ locations to describe the mobility patterns and spatial distance between locations.
The details are as follows:

\textbullet\  \textit{Courier Mobility Matrix ($\mathbf{M_C^{\Delta t}} \in \mathbb{R}^{\Delta t \times N \times N}$)}: Each element in the matrix is the number of transitions between each pair of locations during time slot $\Delta t$ in the historical records.

\textbullet\  \textit{Spatial Distance Matrix ($\mathbf{M}_D \in \mathbb{R}^{N \times N}$)}: Each element in the matrix represents the spatial distance between each pair of locations.

Next, to benefit computational efficiency, %at each step, 
%available information about the package locations,
for each pending package set $F_t$ consisting of $L_f$ packages where $t \in \Delta t$, 
%based on the locations of the $L_f$ pending packages, 
we select the corresponding rows and columns based on their locations, to form the matrices $\mathbf{M}_C^{'}, \mathbf{M}_D^{'} \in \mathbb{R}^{L_f \times L_f}$ from $\mathbf{M_C^{\Delta t}}, \mathbf{M}_D$ respectively, and normalize them by the row.
Then the two matrices are used to weight the previous intermediate probability $u_j$ at each step %produced by the last component 
by matrix multiplication:
\begin{equation}
    [\mathbf{R}_C; \mathbf{R}_D]= \mathbf{u}_j \times \sigma(Linear([\mathbf{M}_C^{'}; \mathbf{M}_D^{'}])),
\end{equation}
where $\sigma$ means the sigmoid function to ensure weights remain positive.
Next, we fuse this information to obtain the final predicted probabilities of packages at step $j$:
%其实下面这里也有mask，但不知道怎么用公式写出来，就先不搞了
\begin{equation}
    P(\pi_{j})=Softmax(\mathbf{u}_j + \mathbf{R}_C + \mathbf{R}_D),
\end{equation}
where $P(\pi_{j})$ is the vector with its element $P(\pi_{j})^i$ representing the probability of package $i$. 
The maximum element is the predicted package $\pi_j$ at the current step:
\begin{equation}
    \pi_{j}=\text{argmax}_i{P(\pi_{j})^i}.
\end{equation}
where $i$ denotes the index of each package. 
Finally, the finished time of $\pi_j$ is estimated with $\mathbf{e}_j$, the output of the LSTM layer which is concatenated with the corresponding representation of the package in $A_t$ at each step via a fully connected layer:
\begin{equation}
    \tilde{y}_j=ReLU(Linear(\mathbf{e}_j||\mathbf{A^{\pi_j}_t})).
\end{equation}

\subsection{Optimization}
%在related里有些工作也用了aux，如compaeta
For the main time prediction task, we employ the MSE as the main loss function:
\begin{equation}
%\mathcal{L}_{main} = - \sum_{t \in T} \sum_{o \in O_t^d} {(t_o- \tilde{t}_o)^2 },
\mathcal{L}_{main} = - \frac{1}{|\mathcal{D}|} \sum_\mathcal{D} \sum_{j=1}^{n_d'} {({y}_j- \tilde{{y}}_j)^2},
\end{equation}
%where $O_t^d$ denotes the \textbf{delivery} pending packages at $t$ since the pickup is not involved in the time prediction task, $t_o$ is the ground truth of delivery time of package $o$, and  $P\tilde{t}_o$ is the predicted delivery time of $o$ produced by the model.
where $n_d'$ denotes the number of pending delivery packages in each sample, and ${y}_j$ and $\tilde{{y}}_j$ are the real value and estimated time of delivery generated in Section ~\ref{Mobility component} with pickup being masked in the loss calculation.

For the auxiliary route prediction task, we choose the cross entropy as the auxiliary loss function which is widely used in multi-label classification, formulated as:
\begin{equation}
\mathcal{L}_{aux} = - \frac{1}{|\mathcal{D}|} \sum_\mathcal{D} \sum_{j=1}^{n'} \log {P(\pi_j^{true})},
\end{equation}
%where $T$ denotes the overall training periods, $O_t$ denotes all the pending packages set including both delivery and pickup at $t$, $y_o$ is the order of package $o$ in the courier's route, and  $P(y_o|\theta)$ is the predicted probability of $o$ produced by the model.
where $\mathcal{D}$ denotes the training dataset, 
$n'=n_d'+n_p'$ denotes the number of pending packages including both delivery and pickup in each sample, 
$\pi_i^{true}$ is the actual package courier completes at step $j$, and $P(\cdot)$ is the final predicted probability generated by Section ~\ref{Mobility component}.

Since the two loss functions differ in scales, the final loss is the weighted combination of the two losses above, \emph{i.e.}, $\mathcal{L}=\mathcal{L}_{main}+\alpha*\mathcal{L}_{aux}$, where $\alpha$ is a learnable parameter.

\section{EXPERIMENTS}
\subsection{Experimental Setup}
\subsubsection{Dataset}\label{dataset overview}
%更改下面的数据
As summarized in Table~\ref{Dataset Summary}, the two real-world datasets provided by JD Logistics includes the daily records of couriers in Beijing and Tianjin, respectively. 
In each dataset, the number of pickup samples is less than 10\% of the delivery.
%171 couriers in Beijing, from February 1st, 2023, to July 31st, 2023. In the dataset, the pickup volume is less than 10\% of the delivery volume. %As each courier has a record for 286 days, the total number of samples is 155584.
Each dataset is divided into training, validation and testing according to date in the ratio of 6:2:2. 
We also collect data from January 1st to February 28th, 2023 to calculate transitions between AOIs (Area of Interest) in the mobility prediction component. 

% % % \vspace{-4mm}
% \begin{table}[htbp]
% %\scalebox{0.1}
% \centering
% \caption{The statistics of our dataset.}\label{Dataset Summary}
% % % \vspace{-4mm}
% \resizebox{1\linewidth}{!}{
% \begin{tabular}{|c|c|c|c|}
% \hline
% \textbf{} &  \textbf{Training} & \textbf{Validation} & \textbf{Testing}\\
% \hline
% \textbf{\# of deliveries} &  1633978 & 638063 & 549517\\
% \textbf{\# of pickups} &  134992 & 53672 & 40564\\
% \textbf{\# of couriers} &  171 & 171 & 171\\
% \textbf{\# of AOIs} &  952 & 952 & 952\\
% \textbf{\# of days} &  1, Mar. - 31, May. & 1, June. - 30, Jun. & 1, Jul. - 31, Jul.\\
% \hline
% \end{tabular}
% }
% \end{table}
% % % \vspace{-4mm}

\vspace{-4mm}
\begin{table}[htbp]
%\scalebox{0.1}
\centering
\caption{The statistics of our dataset.}\label{Dataset Summary}
% \vspace{-2mm}
\resizebox{1\linewidth}{!}{
\begin{tabular}{|c|c|c|c|c|c|}
\hline
\textbf{Dataset}& \textbf{Date} &  \textbf{Deliveries} & \textbf{Pickups} & \textbf{Couriers}& \textbf{AOIs}\\
\hline
\textbf{Beijing} &1, Mar. - 31, Jul.,2023& 2821558& 229228 & 171 & 952\\
\textbf{Tianjin} &1, Aug. - 31, Dec.,2023& 2109782& 165846 & 129 & 691\\
%2128543 & 172926 & 129 & 691
\hline
\end{tabular}
}
\end{table}
\vspace{-4mm}

%As show in Table~\ref{datasets}, o
In detail, the dataset is comprised of five kinds of information: logistics information, geographical information, weather forecast information, date information and courier information. 
The logistics information records detailed information for each package, including package type (delivery or pickup), package assignment time to couriers, promised time, time slot of the promised time, whether overdue, actual finish time, weight, volume, and latitude-longitude coordinates based on its address via Geocoding ~\cite{geocoding}. 
%~\footnote{https://en.wikipedia.org/wiki/Address\_geocoding}.
%Remarkably, features of timestamp $t$ contain the timely rate of $t-1$ when predicting the timely rate of $t$. 
Additionally, the remaining completion time, as well as the spatial distance from couriers' current location to pending packages, will also be calculated at each time step as part of the model's input. 
Moreover, packages with the same courier, customer, AOI and time are aggregated together to avoid duplication~\cite{wen2023modeling}.
The geographical information contains the 45-bit Geohash code of stay points, which is detected from couriers' historical trajectories using DTInf ~\cite{ruan2020doing,hong2023autobuild} as AOIs, and each package will be matched to the closest stay point.
%the boundary of road districts and the road network of the corresponding urban area from OpenStreetMap with the parameter "network type" set to "drive", indicating that the obtained roads are passable by vehicles.
The regional weather forecast information 
%in the corresponding area 
is acquired from ~\cite{lishitianqi}
%~\footnote{http://lishi.tianqi.com/}, 
including daily weather type forecasting, predicted highest, lowest and average temperatures. 
The date type information includes 
%two types of features: 
the day of the week and whether the day is a holiday.
The courier information contains courier profiles such as gender, length of employment and age.
%The epidemic information is obtained from ~\footnote{https://news.sina.cn/zt\_d/yiqing0121}.

In addition, couriers' decision making will be affected by newly dispatched packages. Thus, when constructing samples, we need to split the delivery routes into samples accordingly, which can be found in Appendix ~\ref{Newly Dispatched}.

% In addition, couriers' decision making will be affected by newly dispatched packages. 
% In the real world, delivery packages are pre-allocated to the courier before he/she commences his/her daily work, while pickup packages are assigned continually throughout the day. 
% Whenever a pickup package is dispatched, the courier needs to re-plan the route, which will affect the pending packages consequently. 
% Therefore, it is necessary to consider this issue when constructing the dataset.
% As shown in Figure~\ref{split}, suppose that the courier has been dispatched delivery of $\{o_1,o_2,o_3,o_5,o_6\}$ at the beginning. After delivering $\{o_1,o_2,o_3\}$, the pickup of $o_4$ is dispatched to the courier. Thus he/she turns to pick up $o_4$ first, which results in the change in the sequence of order and time of $\{o_5,o_6\}$. 
% To mitigate the impact of such unexpected dispatched packages, we split the original package route into two at $o_4$ when constructing datasets. 

% % \vspace{-3mm}
% \begin{figure}[htbp]
% \centering
% \includegraphics[width=\linewidth]{figures/split.pdf}
% % \vspace{-5mm}
% \caption{An example of splitting route at the newly dispatched package.
% } 
% \label{split}
% % \vspace{-6mm}
% \end{figure}

\subsubsection{Baselines}
We choose the following methods as baselines, 
which are commonly adopted by industry: historical average delivery time (AVG), 
Extreme Gradient Boosting (XGB)~\cite{chen2016xgboost} and Random Forest (RF),  %Linear regression (LR) 
are conventional machine learning methods. 
MLP, OFCT~\cite{zhu2020order}, the Bidirectional Long short-term memory (BiLSTM), 
%which was deployed online previously in JD Logistics, 
%Transformer, 
DeepETA~\cite{wu2019deepeta}, FDNET~\cite{gao2021deep}, DeepRoute~\cite{wen2021package}, 
%\text{$M^2G4RTP$}
RankETPA~\cite{wen2023enough}, 
CP-Route~\cite{wen2023modeling},
M\textsuperscript{2}G4RTP~\cite{cai2023m}, 
are deep learning-based methods.
Excluding the Geohash code which is difficult to introduce in traditional machine learning models, the features utilized in various models remain consistent. 
For baselines like DeepRoute, M\textsuperscript{2}G4RTP and CP-Route which only predict the arrival route of packages, 
%The typical approach for route prediction involves using the attention-based recurrent neural network to predict the sequence. Building upon this, the method for both route prediction and time prediction like RankETPA [3], usually utilizes the output of the aforementioned recurrent neural network as inputs to a fully connected layer, thus obtaining the predicted time. Following this approach, 
we follow the common approach of using the output of the attention-based recurrent neural network as inputs of the fully connected layer to estimate the arrival time~\cite{wen2023enough}. 
% % \vspace{-1mm}
\subsubsection{Metrics}
For the time prediction task, the Rooted Mean Absolute Percentage Error (RMSE) which quantifies the average absolute difference between the prediction and the ground truth, and the Mean Absolute Percentage Error (MAPE) which is commonly used in the ETA task to represent the deviation between the predicted result and the true value, are used as the evaluation metrics formulated as follows: 
\begin{equation}
\text{RMSE} = \sqrt{\frac{1}{N}\sum_{i=1}^{N}(y_i - \tilde{y}_i)^2},
\end{equation}
\begin{equation}
\text{MAPE} = \frac{1}{N}\sum_{i=1}^{N}\left|\frac{y_i - \tilde{y}_i}{y_i}\right| \times 100\%,
\end{equation}
where $y_i$ and $\tilde{y}_i$ are the actual delivery time and the predicted delivery time, respectively. $N$ denotes the number of pending tasks.

To evaluate the route prediction task, we use two metrics, the Location Mean Deviation (LMD) which calculates the deviation between prediction and ground truth, and the Hit-Rate@k which measure the similarity between two sequences, are used as the evaluation metrics formulated as follows: 
\begin{equation}
\text{LMD} = \frac{1}{N} \sum_{i=1}^{N} \left| \pi_i - \tilde{\pi}_i \right|,
\end{equation}
\begin{equation}
\text{HR@k} = \frac{ \pi_{[1:k]} \cap \tilde{\pi}_{[1:k]} }{k}, 
\end{equation}
where $\pi_i$ and $\tilde{\pi}_i$ are the actual delivery package and the predicted delivery package, respectively. $N$ denotes the number of pending tasks.

\subsubsection{Model Setting}
The batch size of each epoch is set to 128. 
The package lengths $L_h$ and $L_f$ are set to 15. 
Pending package sets with length less than $L_f$ will be zero padded, and for those with lengths greater than $L_f$, only the latest $L_f$ completed packages will be reserved. 
In the package spatio-temporal correlation extraction module, we stack two Transformer encoder blocks with four multi-head attention in each block, and the hidden sizes $d_{model}$ is set to 64.
%这里有必要说嘛？两层encoder是不是不好？
In the pickup influencing pattern learning module, the number of stored patterns $L_m$ equals 20. 
We also apply a dropout layer with a dropout rate of 0.1 to the LSTM. % and FC layers. 
%Our model is optimized using Adam and trained with 100 epochs as the performance can converge early 
We optimize the model with Adam in an end-to-end manner with the learning rate set to 0.0001.
We also calculate the inference time per batch.

% \vspace{-4mm}

\subsection{Performance Comparison} 
We evaluate each model for five independent rounds, and take the average of five rounds for each metric. 

\begin{table*}[t]
    \centering
    \caption{Performance of our model and baselines.}\label{performance}
    
    \resizebox{\textwidth}{!}{
    \begin{tabular}{c|c|c|c|c|c|c|c|c|c|c}
        \hline
        \multirow{2}{*}{\textbf{Model}} & \multicolumn{5}{c|}{\textbf{Beijing Dataset}} & \multicolumn{5}{c}{\textbf{Tianjin Dataset}} \\
        \cline{2-11}
        & \textbf{RMSE (min)} & \textbf{MAPE (\%)} & \textbf{LMD} & \textbf{HR@3 (\%)} & \textbf{Inference time (s)} & \textbf{RMSE (min)} & \textbf{MAPE (\%)} & \textbf{LMD} & \textbf{HR@3 (\%)} & \textbf{Inference time (s)} \\
        \hline
        AVG & 34.95 & 113.62 & - & - & - & 35.23 & 114.78 & - & - & - \\
        RF & 29.38 & 86.25 & - & - & 0.13 & 29.75 & 87.02 & - & - & 0.13 \\
        XGB & 28.06 & 86.01 & - & - & 0.02 & 28.44 & 86.50 & - & - & 0.02 \\
        
        MLP & 28.77 & 82.76 & - & - & 0.01 & 28.82 & 83.15 & - & - & 0.01 \\
        OFCT & 28.56 & 70.93 & - & - & 0.29 & 28.75 & 71.30 & - & - & 0.30 \\
        BiLSTM & 23.45 & 32.84 & - & - & 3.72 & 23.86 & 33.25 & - & - & 3.71 \\
        DeepETA & 17.14 & 27.90 & - & - & 5.78 & 17.45 & 28.34 & - & - & 5.79 \\
        \hline
        FDNET & 19.16 & 26.89 & 2.41 & 54.01 & 3.92 & 21.53	&31.18	&2.42&54.20&	 3.92 \\
        DeepRoute & 16.09 & 25.02 & 2.40 & 54.75 & 4.05 & 15.89	&26.94&2.41 &54.34&	 4.07 \\
        RankETPA & 15.98 & 24.87 & 2.38 & 55.02 & 4.03 & 15.23&	24.34&	2.39 &54.95	 & 4.04 \\
        M\textsuperscript{2}G4RTP & 15.83 & 23.28 & 2.36 & 55.47 & 6.02 & \underline{15.07}&	23.97& 2.18	&\underline{56.66}	 & 6.11 \\
        CP-Route & \underline{15.25} & \underline{23.01} & \underline{2.32} & \underline{56.22} & 3.97 & 15.09 &\underline{23.83} & \underline{2.17}	 &56.41& 3.99 \\
        \hline
        TransPDT & $\mathbf{14.71}$ & $\mathbf{18.86}$ & $\mathbf{2.27}$ & $\mathbf{56.79}$ & 4.12 & $\mathbf{14.62}$ & $\mathbf{22.47}$ & $\mathbf{2.12}$ & $\mathbf{57.39}$ & 4.10 \\
        \hline
        Improvement &3.54\%  & 18.04\% & 2.16\% & 1.01\% & - & 3.12\% & 5.71\% & 2.30\% & 1.74\% & -  \\
        \hline
    \end{tabular}
    }
\end{table*}

As can be observed in Table~\ref{performance}, 
the AVG method performs the worst in both metrics, 
and the performance of deep learning-based approaches is generally superior to that of conventional methods basically, %since deep learning-based methods are more capable of capturing the temporal dependencies in the sequence.
showing their capability of modeling complex influential factors of package delivery time.

%Notably, the AVG method performs the worst in both metrics, demonstrating the method exhibits significant deviation and can not be applied directly.

Additionally, among deep learning-based methods, MLP and OFCT perform relatively worse, due to the inability to capture the spatial-temporal correlations among packages.
%among conventional methods, linear regression performs the worst, suggesting the existence of complex nonlinear relationships in features. 
Furthermore, the performance of both DeepETA and FDNET is comparable but both are inferior to DeepRoute and RankEPTA. This is because, compared to the latter, DeepETA is unable to forecast the delivery route of packages, and FDNet fails to capture the correlations between pending packages.

Moreover, CP-Route exhibits the best performance among all baselines, since it integrates couriers' mobility patterns between locations into prediction. 
%Specially, it outperforms M\textsuperscript{2}G4RTP, indicating that 
Nonetheless, it performs worse than our model, as it fails to capture the complicated interdependencies between pickup and delivery.

%among deep learning-based methods, DNN performs worst due to the lack of modeling non-linear relationships and spatial-temporal dependencies. 
%Moreover, RankETPA shows better performance than DeepRoute 
In general, our model achieves the best performance in all metrics. 
In the main task of time prediction, TransPDT outperforms its best competitors
by 3.54\% in RMSE, 18.04\% in MAPE on Beijing Dataset; 3.12\% in RMSE, and 5.71\% in MAPE on Tianjin Dataset. 
Besides, in the auxiliary route prediction task, our model outperforms all baselines by at least 2.16\% in LMD, 1.01\% in HR@3 on Beijing Dataset; 2.30\% in LMD, 1.71\% in HR@3 on Tianjin Dataset, indicate that more accurate road prediction can enhance model's performance on time prediction.
Additionally, our model's inference time is comparable to other deep learning-based methods. 
Remarkably, none of the deep learning-based methods perform as well as TransPDT. 
This is because none of the aforementioned methods model the courier mobility preferences effectively, and fail to model the impact of pickup on delivery with insufficient pickup samples.%, while our model models the correlation between couriers and between road districts additionally and models the anomaly events separately.

% \vspace{-3mm}

% \vspace{-2mm}
\begin{figure}[htb]
\includegraphics[width=\linewidth]{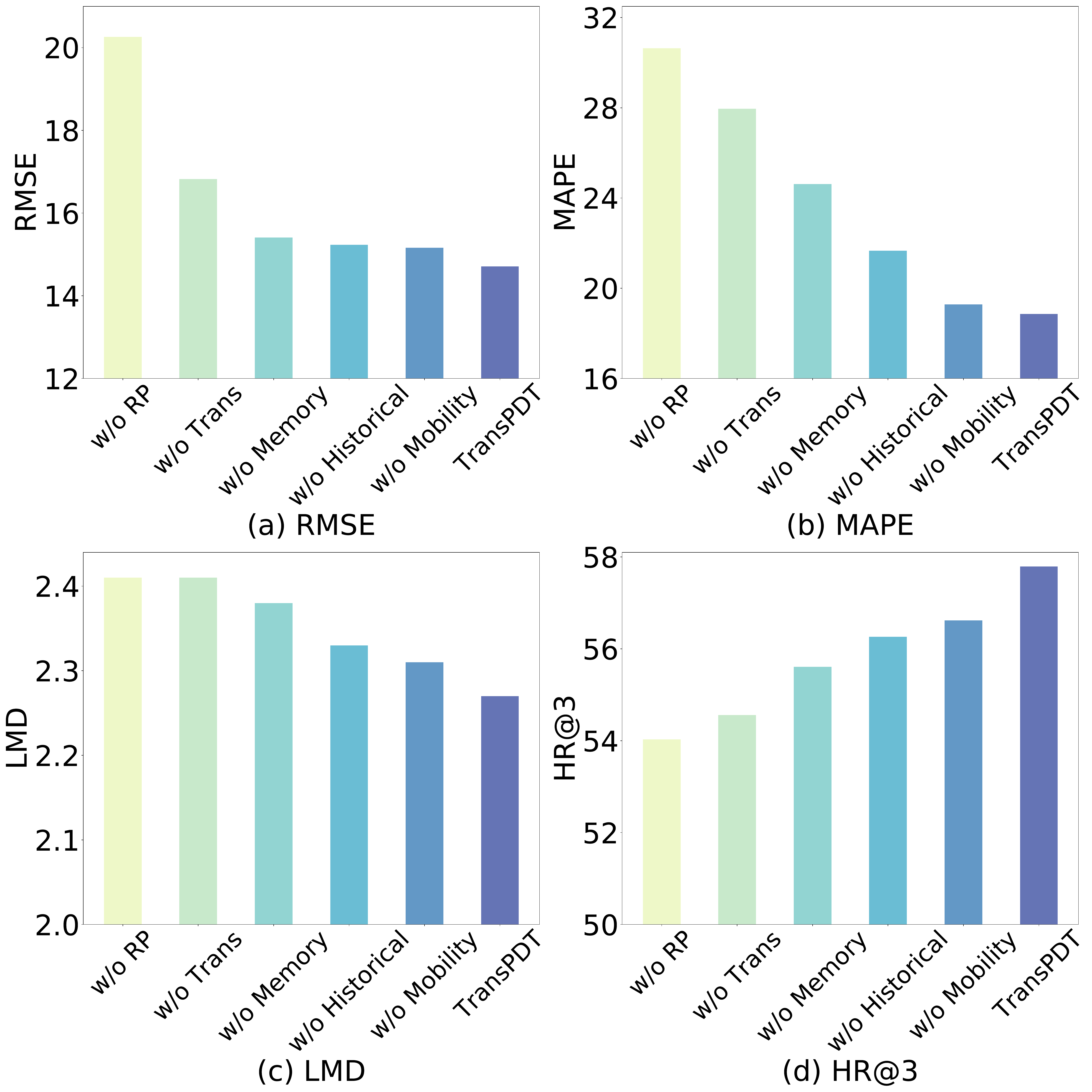}
% \vspace{-6mm}
\caption{Ablation study.} \label{ablation_new}
% \vspace{-4mm}
\end{figure}
% \vspace{-4mm}

\subsection{Ablation Study}
%1.移除两个Transformer：应该和fdnet差不多略好
%2.移除memory：已实现，应该比ranketpa好
%3.移除两个矩阵：已实现，应该比ranketpa好
%4.移除排序网络但有memory：和DeepETA应该差不多略好
We introduce several variants of the model to validate the efficiency of each component. %, i.e., 
To be specific, the two matrices in the mobility prediction component  are not incorporated, 
%and use the rest features as the input for Courier spatial learning, which is 
denoted as "w/o Mobility". 
In the same way, 
"w/o Trans" removes the Transformer encoder of the package spatio-temporal correlation extraction module,
"w/o Historical" removes the historical package route branch in the package spatio-temporal correlation extraction module,
"w/o Memory" removes the pattern memory of the pickup influencing pattern learning module,
and "w/o RP" removes the whole route prediction procedure of the mobility-based package prediction module, respectively.
%we also remove the GCN network in Courier spatial learning, the LSTM network in Courier spatial learning, the RNN network in Anomaly learning, the memory network in  with corresponding models named after "w/o GCN", "w/o LSTM", respectively. 

As shown in Figure~\ref{ablation_new}
%Table~\ref{Ablation}, 
the TransPDT outperforms the best competitors by 2.97\% in RMSE, 2.18\% in MAPE, 1.73\% in LMD and 2.07\% in HR@3 on Beijing Dataset. The absence of any component will lead to deterioration in the overall performance obviously. 
This demonstrates that
%the validity of the design of each module and component in xx, demonstrating that the four types of information, \emph{i.e.}, the correlation between road districts, the correlation between couriers, the sequential dependencies and anomaly information, can contribute to the prediction performance.
the spatio-temporal relations between packages will determine the delivery time on a large scale, 
courier's historical travel behaviors will affect future transition, 
the newly dispatched pickup tasks will affect the couriers' delivery process, 
incorporating couriers' preferences can enhance the model's performance, 
and courier's travel route has a significant impact on the delivery time of packages.

% % \vspace{-3mm}
% %[add results here! CHECK font size?]
% \begin{table}[htbp]
% \centering
% \caption{Ablation study results.}\label{Ablation}
% % \vspace{-2mm}
% \begin{tabular}{|c|c|c|}
% % \begin{tabular}{m{6cm}<{\centering}|m{1cm}<{\centering}|m{1cm}<{\centering}|m{1cm}<{\centering}}
% \hline
% \textbf{Model} &  \textbf{RMSE} & \textbf{MAPE} \\
% \hline
% {w/o Transformer} &  16.82 & 27.96 \\
% {w/o Historical} &  15.23 & 21.66 \\
% {w/o Memory} &  15.41 & 24.62\\
% {w/o Mobility} &  15.16 & 19.28 \\
% {w/o RP} &  20.26 & 30.64\\
% % {w/o RNN} &  0.1785 & 0.0549 \\
% % {w/o Memory+RNN} &  0.1819 & 0.0568\\
% {TransPDT} &  $\mathbf {14.71}$ & $\mathbf {18.86}$\\
% \hline
% \end{tabular}
% % \vspace{-2mm}
% \end{table}
% % \vspace{-4mm}

% % \vspace{-1mm}
\begin{figure*}[htb]
\includegraphics[width=\linewidth]{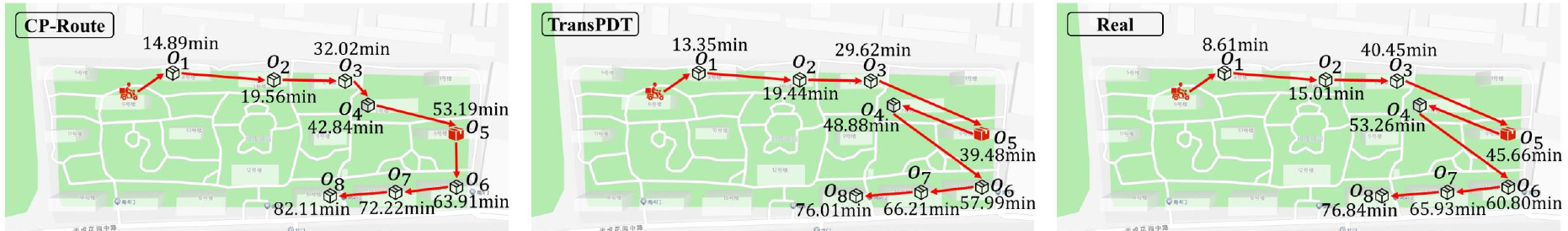}
\vspace{-6mm}
\caption{ An actual case study.} 
\vspace{-2mm}
\label{cases}
\end{figure*}
\vspace{-3mm}

\subsection{Hyper-parameter \& Case study}
We explore the impact of two important hyper-parameters of $L_m$, the number of patterns stored in the pickup pattern attentional network and $L_f$, the length of pending package sets. 
During the experiment, while changing one parameter, other parameters will be kept constant. The results on Beijing dataset 
% are shown in Figure~\ref{Tprevious} 
can be found in ~\ref{Hyper}
which demonstrates the significance of patterns and validates the robustness and stability of our method.

\begin{figure}[htb]
\includegraphics[width=\linewidth]{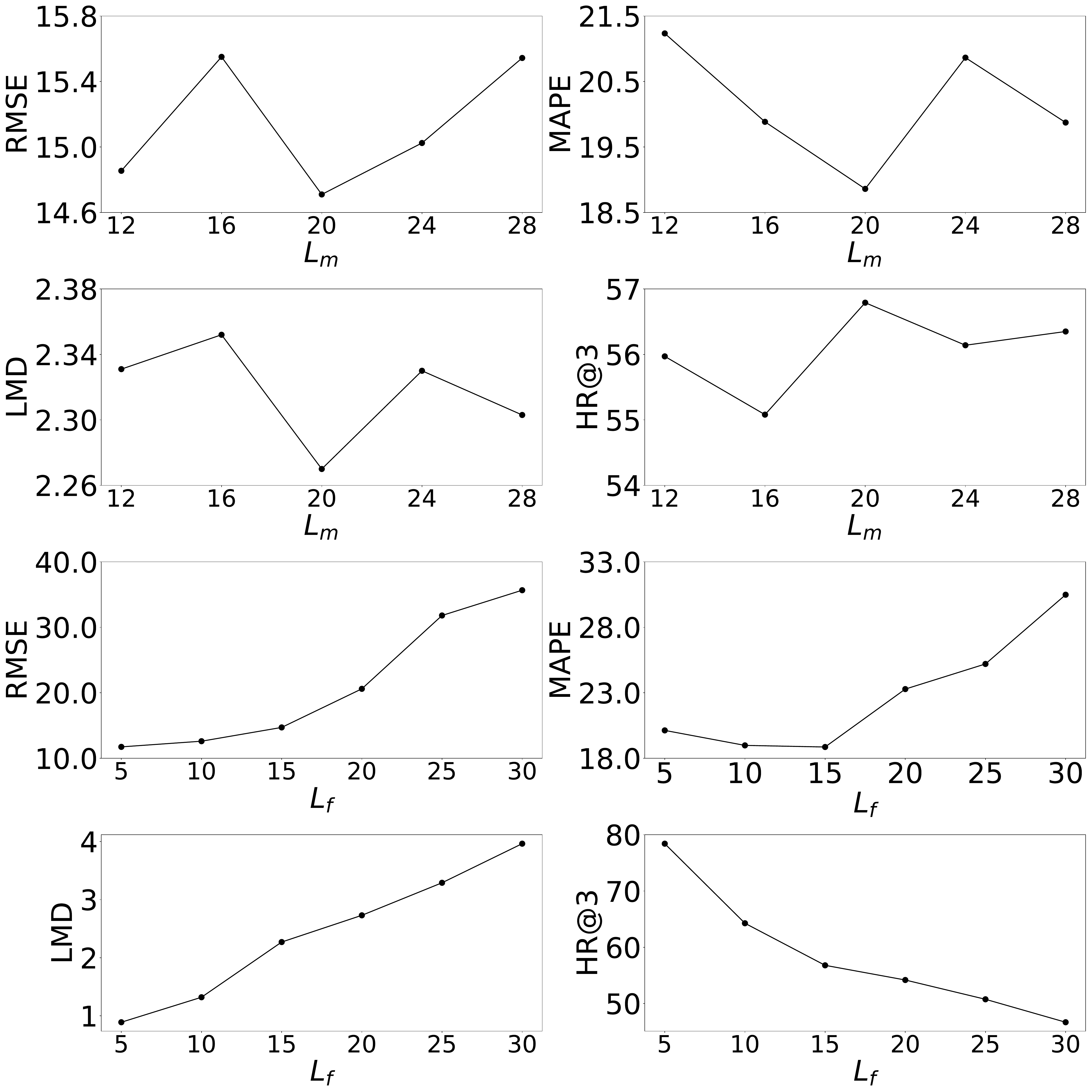}
\vspace{-8mm}
\caption{The impact of hyper-parameters.} \label{Tprevious}
\vspace{-4mm}
\end{figure}
% \vspace{-2mm}

We first test the impact of $L_m$ which controls the number of stored patterns by varying it from 12 to 28 at an interval of 4. 
As can be observed, our model achieves the best performance in all metrics when $L_m = 24$. 
The possible reason could be that a small $L_m$ fails to learn the relations between pickup and delivery sufficiently, while a large $L_m$ may introduce noisy information.
%Specially, the results across various $L_m$ all outperform the best baseline, verifying the significance of exploring the impact of pickup on delivery.
%indicating that the model's  performance will be impacted when input information is insufficient or redundant.

We also modify the pending set length $L_f$ ranging from 5 to 30 at an interval of 5. 
From the graph, we can observe that our model's performance remains relatively stable when $L_f$ does not exceed 20. 
However, RMSE and MAPE increase gradually as $L_f$ increases beyond 20, yet our method still makes promising results.
Note that the best MAPE is not achieved when $L_f = 5$. This is because couriers make route decisions based on spatio-temporal constraints of all pending packages, while a smaller $L_f$ will result in information loss of pending packages.
In contrast, as $L_f$ increases, not only will it become more challenging to predict the delivery sequence of pending packages, but also couriers may engage in activities such as rest and meals instead of maintaining continuous work conditions facing multiple tasks.
%We can observe that the optimal performance is achieved when $L_m = 12$, suggesting that lacking or excessive external memory will lower model performance.

%\subsection{Case Study}
We also conduct a case study to validate the superiority of our model in learning the impact of limited pickup samples on couriers' delivery decision making. 
As shown in Figure~\ref{cases}, the courier delivers packages in ascending order of spatial proximity. After delivering $o_3$, he does not proceed to deliver the closer package $o_4$. Instead, he moves farther away to pick up package $o_5$. 
CP-Route mainly captures the majority delivery sample patterns, while TransPDT also has the ability to learn the influence of minority pickup samples on couriers' delivery behavior.
In this case, the MAPE of the sample is TransPDT 17.41\% and CP-Route 22.71\%, 
RMSE is TransPDT 5.30 and CP-Route 6.83, 
and LMD is TransPDT 0.0 and CP-Route 0.25, 
indicating that our method is capable of capturing the patterns of limited pickup on couriers' delivery process. 
%We show that TransPDT is capable to learn the impact of limited pickup samples on couriers' delivery decision making.
% As shown in Figure~\ref{cases}, the courier delivers packages in ascending order of spatial proximity. After delivering $o_3$, he does not proceed to deliver the closer package $o_4$. Instead, he moves farther away to pick up package $o_5$. 
% CP-Route mainly captures the majority delivery samples patterns, while TransPDT also has the ability to learn the influence of minoirty pickup samples on couriers' delivery behavior.
% In this case, the MAPE of the sample is TransPDT 17.41 and CP-Route 22.71, and RMSE is TransPDT 5.30 and CP-Route 6.83.

\section{Deployment}
\subsection{Deployment System}
A package delivery time prediction system based on TransPDT is deployed internally in JD Logistics. It currently provides convenience for logistics station managers in tracking the work status of couriers and ensuring on-time delivery.
As shown in Figure~\ref{deploy illustration}, the system consists of the following three layers:
%\emph{i.e.}, the offline training layer, the online inference layer and the interface layer.

\begin{figure}[htbp]
\centering
\includegraphics[width=\linewidth]{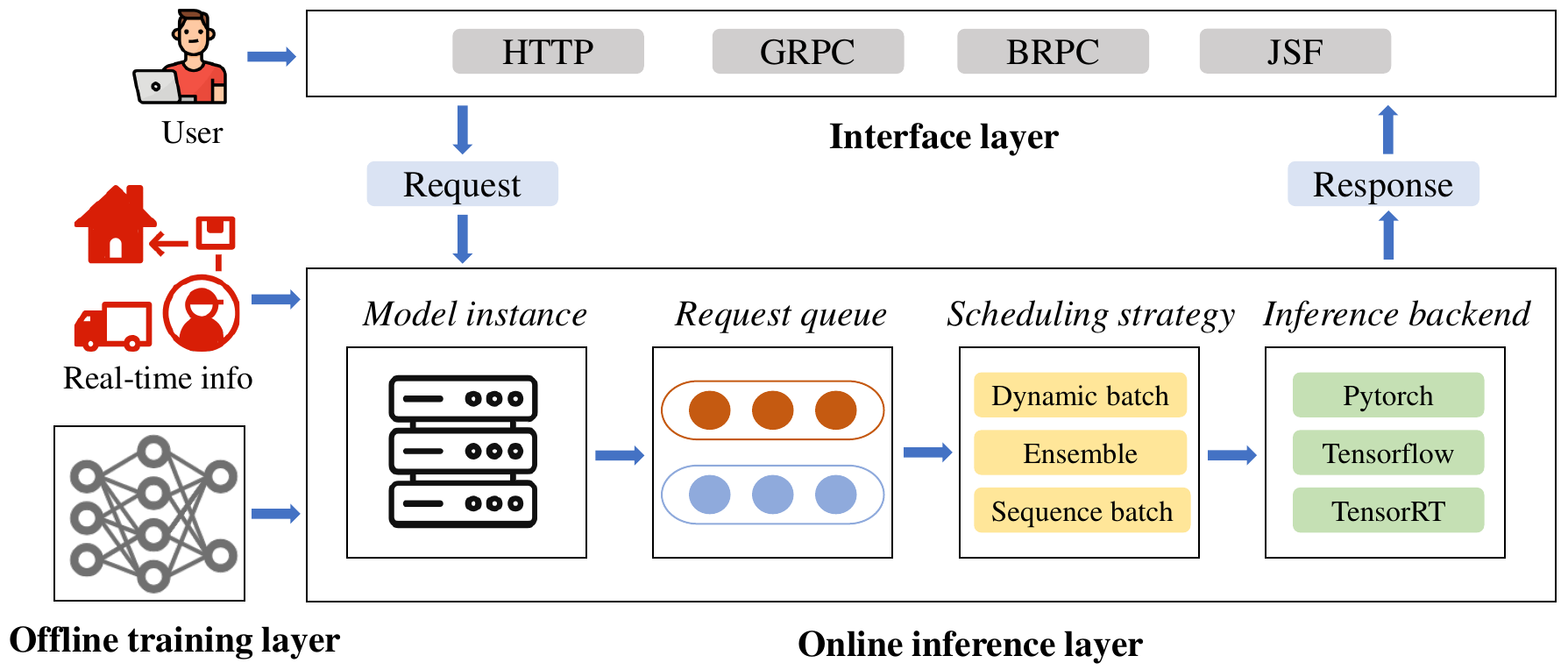}
\vspace{-6mm}
\caption{Deployment system overview.} 
\label{deploy illustration}
% \vspace{-5mm}
\end{figure}
\vspace{-3mm}

\begin{itemize}[leftmargin=*]
\item \textbf{Offline training layer:} In this layer, we train the model offline with Pytorch 1.21.1 in Python 3.7 environment with Intel(R) Xeon(R) CPU E5-2683 v4 @ 2.10GHz (CPU) and one NVIDIA Tesla P40 (GPU).
\item \textbf{Online inference layer:} This layer is divided into four components. The pre-trained model is deployed in the model instance component, which is a server cluster of machines with Intel(R) Xeon(R) CPU E5-2683 v4 @ 2.10GHz, 4 cores, 16 GB RAM and NVIDIA Tesla P40.
The request queue component implements the asynchronous request processing through request queue.
The scheduling strategy component utilizes various scheduling strategies, such as the dynamic batch, ensemble and sequence batch, to handle requests.
The inference backend component supports multiple inference backends for a given task, like Pytorch, Tensorflow and TensorRT.
\item \textbf{Interface layer:}, 
Once the user initiates a request, 
%the interface layer serves as the entry point for communication within the algorithm platform.
the interface layer handles the standardization of communication protocols such as Baidu Remote Procedure Call (BRPC), Google RPC (GRPC), HyperText Transfer Protocol (HTTP) and JavaServer Faces (JSF), and transmits it seamlessly to the inference layer, which will then return the predicted delivery time of each package as the response with real-time information.
\end{itemize}

% In the real world, after the courier accepts all the packages to be delivered in the logistics station, his/her current information %will be sent as query input to the deployed model, which will return the  to the station manager. 
% will be resolved to the deployed model. 
% Whenever the courier completes a package or is dispatched a new package, the model will update prediction results based on the latest information.

\vspace{-4mm}
\begin{figure}[htp]%[t!]
\centering
\includegraphics[width=\linewidth]{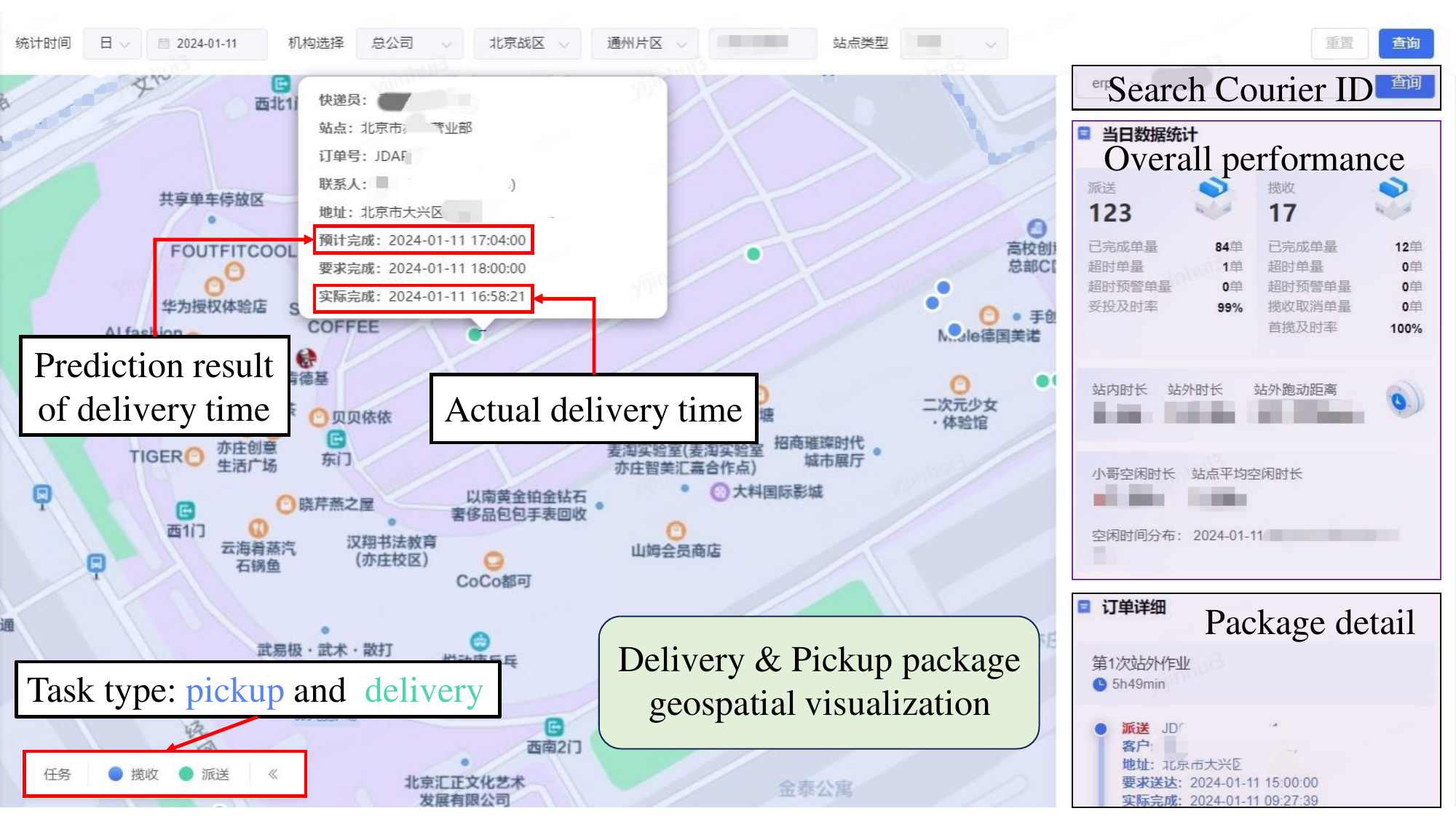}
\vspace{-4mm}
\caption{System interface.} 
\label{deploy}
% \vspace{-6mm}
\end{figure}
\vspace{-6mm}

\subsection{Online Deployment}
The system interface is illustrated in Figure~\ref{deploy}.
On the right side of the interface, after the station manager enters the courier's ID, an overview of the courier's daily work performance will be displayed, including the total number of daily deliveries and pickups, the completed number of different tasks, timely rate which measures the percentage of packages delivered/pickup before promised time~\cite{yi2024rccnet}, packages are about to expire, etc. 
Additionally, the detailed information of each package is also listed below. 
On the left side of the interface is the geographic distribution of packages, where the blue marker denotes the pickup and the green marker denotes delivery. The station manager can click on a package to view its main information, including the responsible courier, logistics station, waybill code, customer, address, predicted delivery time generated by TransPDT, promised delivery time, and actual delivery time. Among them, the actual delivery time will be updated once the courier delivers the package successfully.
%When the station manager identifies orders 
For packages that are about to exceed the promised time, station managers will either remind the responsible couriers via phone, or assign additional crowdsourcing to assist their work, to ensure customers' satisfaction. 
Further, the station manager will reassign pickup tasks from couriers with poor predicted performance to those with better predicted performance.
%if a courier has the historcial records of package timing out in a specific location for multiple times, the station manager will reassign that location to another courier.

\begin{figure}[htbp]
\centering
\includegraphics[width=\linewidth]{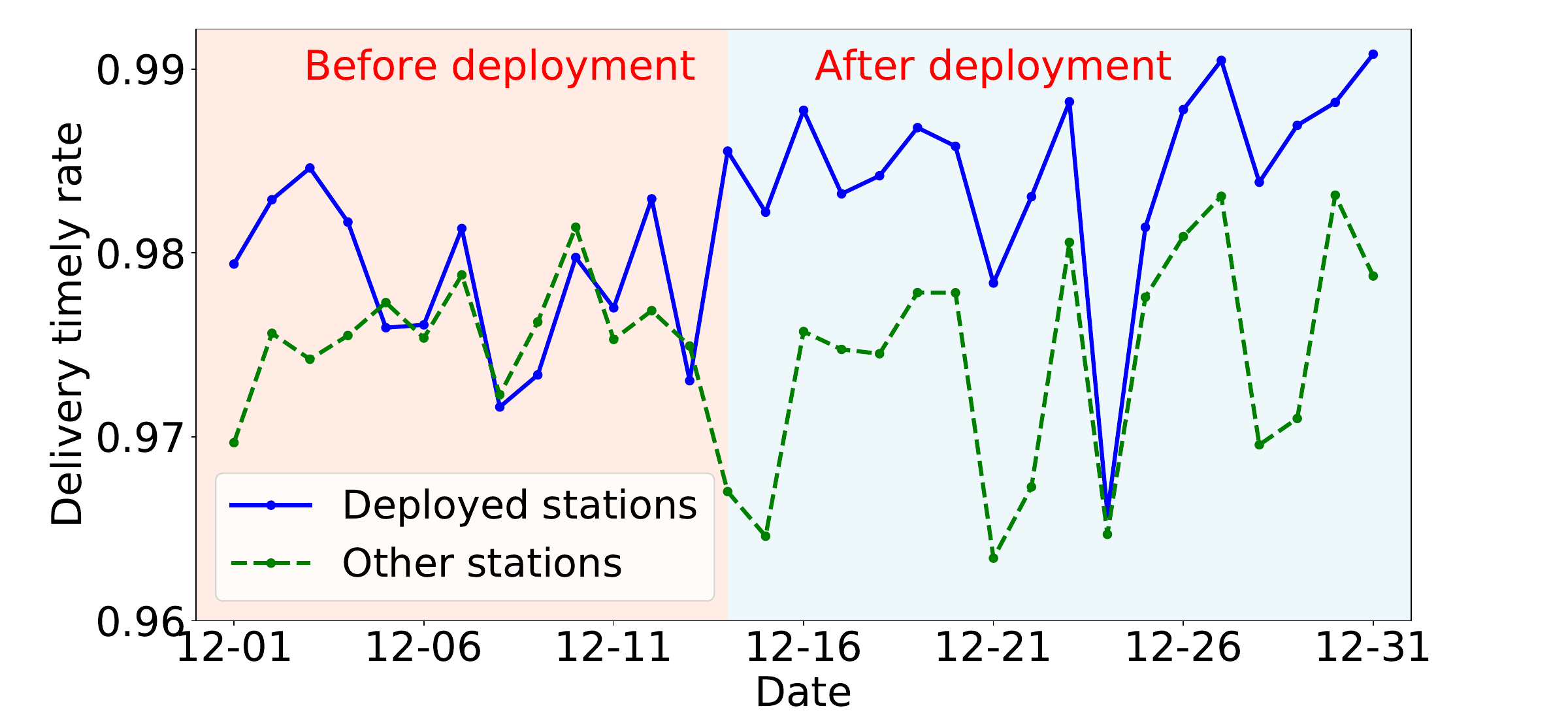}
\vspace{-6mm}
\caption{Delivery timely rate of stations before and after TransPDT deployment.} 
\label{timely}
\vspace{-6mm}
\end{figure}

The system significantly aids station managers in workload allocation and on-time service guarantee. Currently, it is deployed to monitor more than $2,300$ couriers of over 70 logistics stations in Beijing, with millions of requests being handled every day, while there are about 300 non-deployed stations with around $6,000$ couriers in Beijing. 
Further, the company adopts the business metric of delivery timely rate to measure the service quality of couriers, which represents the percentage of packages delivered before promised time each day.
From Figure~\ref{timely}, we observe that before deployment, the average daily delivery timely rate of deployed stations is comparable to that of non-deployed stations. After deployment, the former exceeds the latter by 0.68\%.
Since each courier typically delivers more than $100$ packages daily, this 0.68\% improvement means that more than 0.68 delayed packages can be reduced by our system every day for each courier.
Considering the estimated loss of a delayed package %as well as the conversion rate 
discussed in Section~\ref{intro}, it can be further concluded that, for the $200,000$ couriers of JD Logistics in China, our model has the potential to reduce around $136,000$ delayed packages every day, saving more than $270,000$ RMB (about $38,000$ USD) in daily costs.
% , and generating an additional growth of more than 700 orders for future business.
% The figure of 0.68\% corresponds to one delayed package which can be delivered on-time after system deployment for each courier, 
% since one courier usually delivers more than $100$ packages every day, and the timely rate represents the percentage of on-time delivery packages.
% Consider that the timely rate represents the percentage of on-time delivery packages, and each courier typically delivers more than $100$ packages every day.
% Thus we can conclude this 0.68\% difference corresponds to nearly 1 less delayed delivery package per courier each day after the algorithm deployment.
% It is noteworthy that with respect to the aforementioned estimated loss of a delayed package as well as the conversion rate in Section~\ref{intro},
%is over 20 RMB, and its conversion rate is about 0.4, which implies that for 
%each monitored courier, our model has the potential to save around 20 RMB in costs and bring an additional growth of approximately 0.4 order for future business.
% for the overall current monitored couriers, our model has the potential to save more than $30,000$ RMB (about $4,000$ USD) in daily costs and bring an additional growth of more than 700 orders for future business.
% A larger delivery timely rate indicates that xx, and 1\% improvement in the delivery timely rate will is a markable enhancement to the logistics business, since xx(bring/reduce/increase). 
In the future, we plan to deploy the system on the client side to satisfy customers' experience.

\vspace{-3mm}
% \vspace{-4mm}

% \vspace{-6mm}
\section{Conclusion and Future Work}
In this work, we try to solve the package delivery time estimation in mixed imbalanced delivery and pickup services.
We propose TransPDT, a Transformer-based multi-task 
learning package delivery time prediction framework, which benefits from treating the route prediction as an auxiliary task to boost the performance of time prediction.
The model first employs the Transformer architecture to model the spatio-temporal patterns of couriers' historical routes and future package sets, respectively. 
It then learns and stores the complex patterns of limited pickup samples via the attention mechanism. 
It also treats route prediction as the auxiliary task of time prediction, and incorporates couriers' spatial movement regularities into prediction.
Experiments on real-world data, as well as the internal deployment in JD Logistics, demonstrate the effectiveness of our model, and the model will be further deployed on the client side to serve customers.

We also discuss the limitations of our model. 
Firstly, time constraints like remaining time are directly concatenated with other features as input to the model, which may fail to capture the importance of such information effectively. We plan to design specific modules to process such important information to improve our model's performance.
Secondly, we use couriers' historical behaviors to infer future transition by concatenating the historical representation and future representation together. In the future, we will try to explore how the two types of information interact with each other, and model such interaction to improve the performance of the prediction tasks.

\section{Acknowledgement}
This work was supported in part by the National Natural Science Foundation of China under 62272260, U20B2060, 62171260.

\bibliographystyle{ACM-Reference-Format}
\balance
\bibliography{main}

%%% -*-BibTeX-*-
%%% Do NOT edit. File created by BibTeX with style
%%% ACM-Reference-Format-Journals [18-Jan-2012].

\begin{thebibliography}{45}

%%% ====================================================================
%%% NOTE TO THE USER: you can override these defaults by providing
%%% customized versions of any of these macros before the \bibliography
%%% command.  Each of them MUST provide its own final punctuation,
%%% except for \shownote{}, \showDOI{}, and \showURL{}.  The latter two
%%% do not use final punctuation, in order to avoid confusing it with
%%% the Web address.
%%%
%%% To suppress output of a particular field, define its macro to expand
%%% to an empty string, or better, \unskip, like this:
%%%
%%% \newcommand{\showDOI}[1]{\unskip}   % LaTeX syntax
%%%
%%% \def \showDOI #1{\unskip}           % plain TeX syntax
%%%
%%% ====================================================================

\ifx \showCODEN    \undefined \def \showCODEN     #1{\unskip}     \fi
\ifx \showDOI      \undefined \def \showDOI       #1{#1}\fi
\ifx \showISBNx    \undefined \def \showISBNx     #1{\unskip}     \fi
\ifx \showISBNxiii \undefined \def \showISBNxiii  #1{\unskip}     \fi
\ifx \showISSN     \undefined \def \showISSN      #1{\unskip}     \fi
\ifx \showLCCN     \undefined \def \showLCCN      #1{\unskip}     \fi
\ifx \shownote     \undefined \def \shownote      #1{#1}          \fi
\ifx \showarticletitle \undefined \def \showarticletitle #1{#1}   \fi
\ifx \showURL      \undefined \def \showURL       {\relax}        \fi
% The following commands are used for tagged output and should be
% invisible to TeX
\providecommand\bibfield[2]{#2}
\providecommand\bibinfo[2]{#2}
\providecommand\natexlab[1]{#1}
\providecommand\showeprint[2][]{arXiv:#2}

\bibitem[Cai et~al\mbox{.}(2023)]%
        {cai2023m}
\bibfield{author}{\bibinfo{person}{Tianyue Cai}, \bibinfo{person}{Huaiyu Wan}, \bibinfo{person}{Fan Wu}, \bibinfo{person}{Haomin Wen}, \bibinfo{person}{Shengnan Guo}, \bibinfo{person}{Lixia Wu}, \bibinfo{person}{Haoyuan Hu}, {and} \bibinfo{person}{Youfang Lin}.} \bibinfo{year}{2023}\natexlab{}.
\newblock \showarticletitle{M 2 G4RTP: A Multi-Level and Multi-Task Graph Model for Instant-Logistics Route and Time Joint Prediction}. In \bibinfo{booktitle}{\emph{2023 IEEE 39th International Conference on Data Engineering (ICDE)}}. IEEE, \bibinfo{pages}{3296--3308}.
\newblock


\bibitem[Chen and Guestrin(2016)]%
        {chen2016xgboost}
\bibfield{author}{\bibinfo{person}{Tianqi Chen} {and} \bibinfo{person}{Carlos Guestrin}.} \bibinfo{year}{2016}\natexlab{}.
\newblock \showarticletitle{Xgboost: A scalable tree boosting system}. In \bibinfo{booktitle}{\emph{Proceedings of the 22nd acm sigkdd international conference on knowledge discovery and data mining}}. \bibinfo{pages}{785--794}.
\newblock


\bibitem[Chen et~al\mbox{.}(2022)]%
        {chen2022interpreting}
\bibfield{author}{\bibinfo{person}{Zebin Chen}, \bibinfo{person}{Xiaolin Xiao}, \bibinfo{person}{Yue-Jiao Gong}, \bibinfo{person}{Jun Fang}, \bibinfo{person}{Nan Ma}, \bibinfo{person}{Hua Chai}, {and} \bibinfo{person}{Zhiguang Cao}.} \bibinfo{year}{2022}\natexlab{}.
\newblock \showarticletitle{Interpreting trajectories from multiple views: A hierarchical self-attention network for estimating the time of arrival}. In \bibinfo{booktitle}{\emph{Proceedings of the 28th ACM SIGKDD Conference on Knowledge Discovery and Data Mining}}. \bibinfo{pages}{2771--2779}.
\newblock


\bibitem[de~Araujo and Etemad(2021)]%
        {de2021end}
\bibfield{author}{\bibinfo{person}{Arthur~Cruz de Araujo} {and} \bibinfo{person}{Ali Etemad}.} \bibinfo{year}{2021}\natexlab{}.
\newblock \showarticletitle{End-to-end prediction of parcel delivery time with deep learning for smart-city applications}.
\newblock \bibinfo{journal}{\emph{IEEE Internet of Things Journal}} \bibinfo{volume}{8}, \bibinfo{number}{23} (\bibinfo{year}{2021}), \bibinfo{pages}{17043--17056}.
\newblock


\bibitem[Fang et~al\mbox{.}(2020)]%
        {fang2020constgat}
\bibfield{author}{\bibinfo{person}{Xiaomin Fang}, \bibinfo{person}{Jizhou Huang}, \bibinfo{person}{Fan Wang}, \bibinfo{person}{Lingke Zeng}, \bibinfo{person}{Haijin Liang}, {and} \bibinfo{person}{Haifeng Wang}.} \bibinfo{year}{2020}\natexlab{}.
\newblock \showarticletitle{Constgat: Contextual spatial-temporal graph attention network for travel time estimation at baidu maps}. In \bibinfo{booktitle}{\emph{Proceedings of the 26th ACM SIGKDD International Conference on Knowledge Discovery \& Data Mining}}. \bibinfo{pages}{2697--2705}.
\newblock


\bibitem[Feng et~al\mbox{.}(2018)]%
        {feng2018deepmove}
\bibfield{author}{\bibinfo{person}{Jie Feng}, \bibinfo{person}{Yong Li}, \bibinfo{person}{Chao Zhang}, \bibinfo{person}{Funing Sun}, \bibinfo{person}{Fanchao Meng}, \bibinfo{person}{Ang Guo}, {and} \bibinfo{person}{Depeng Jin}.} \bibinfo{year}{2018}\natexlab{}.
\newblock \showarticletitle{Deepmove: Predicting human mobility with attentional recurrent networks}. In \bibinfo{booktitle}{\emph{Proceedings of the 2018 world wide web conference}}. \bibinfo{pages}{1459--1468}.
\newblock


\bibitem[Feng et~al\mbox{.}(2020)]%
        {feng2020learning}
\bibfield{author}{\bibinfo{person}{Jie Feng}, \bibinfo{person}{Zeyu Yang}, \bibinfo{person}{Fengli Xu}, \bibinfo{person}{Haisu Yu}, \bibinfo{person}{Mudan Wang}, {and} \bibinfo{person}{Yong Li}.} \bibinfo{year}{2020}\natexlab{}.
\newblock \showarticletitle{Learning to simulate human mobility}. In \bibinfo{booktitle}{\emph{Proceedings of the 26th ACM SIGKDD international conference on knowledge discovery \& data mining}}. \bibinfo{pages}{3426--3433}.
\newblock


\bibitem[Feng et~al\mbox{.}(2023)]%
        {feng2023ilroute}
\bibfield{author}{\bibinfo{person}{Tao Feng}, \bibinfo{person}{Huan Yan}, \bibinfo{person}{Huandong Wang}, \bibinfo{person}{Wenzhen Huang}, \bibinfo{person}{Yuyang Han}, \bibinfo{person}{Hongsen Liao}, \bibinfo{person}{Jinghua Hao}, {and} \bibinfo{person}{Yong Li}.} \bibinfo{year}{2023}\natexlab{}.
\newblock \showarticletitle{ILRoute: A Graph-based Imitation Learning Method to Unveil Riders' Routing Strategies in Food Delivery Service}. In \bibinfo{booktitle}{\emph{Proceedings of the 29th ACM SIGKDD Conference on Knowledge Discovery and Data Mining}}. \bibinfo{pages}{4024--4034}.
\newblock


\bibitem[Fu et~al\mbox{.}(2020)]%
        {fu2020compacteta}
\bibfield{author}{\bibinfo{person}{Kun Fu}, \bibinfo{person}{Fanlin Meng}, \bibinfo{person}{Jieping Ye}, {and} \bibinfo{person}{Zheng Wang}.} \bibinfo{year}{2020}\natexlab{}.
\newblock \showarticletitle{CompactETA: A fast inference system for travel time prediction}. In \bibinfo{booktitle}{\emph{Proceedings of the 26th ACM SIGKDD International Conference on Knowledge Discovery \& Data Mining}}. \bibinfo{pages}{3337--3345}.
\newblock


\bibitem[Gao et~al\mbox{.}(2021)]%
        {gao2021deep}
\bibfield{author}{\bibinfo{person}{Chengliang Gao}, \bibinfo{person}{Fan Zhang}, \bibinfo{person}{Guanqun Wu}, \bibinfo{person}{Qiwan Hu}, \bibinfo{person}{Qiang Ru}, \bibinfo{person}{Jinghua Hao}, \bibinfo{person}{Renqing He}, {and} \bibinfo{person}{Zhizhao Sun}.} \bibinfo{year}{2021}\natexlab{}.
\newblock \showarticletitle{A deep learning method for route and time prediction in food delivery service}. In \bibinfo{booktitle}{\emph{Proceedings of the 27th ACM SIGKDD Conference on Knowledge Discovery \& Data Mining}}. \bibinfo{pages}{2879--2889}.
\newblock


\bibitem[Han et~al\mbox{.}(2023)]%
        {han2023ieta}
\bibfield{author}{\bibinfo{person}{Jindong Han}, \bibinfo{person}{Hao Liu}, \bibinfo{person}{Shui Liu}, \bibinfo{person}{Xi Chen}, \bibinfo{person}{Naiqiang Tan}, \bibinfo{person}{Hua Chai}, {and} \bibinfo{person}{Hui Xiong}.} \bibinfo{year}{2023}\natexlab{}.
\newblock \showarticletitle{iETA: A Robust and Scalable Incremental Learning Framework for Time-of-Arrival Estimation}. In \bibinfo{booktitle}{\emph{Proceedings of the 29th ACM SIGKDD Conference on Knowledge Discovery and Data Mining}}. \bibinfo{pages}{4100--4111}.
\newblock


\bibitem[Hong et~al\mbox{.}(2023)]%
        {hong2023autobuild}
\bibfield{author}{\bibinfo{person}{Zhiqing Hong}, \bibinfo{person}{Dongjiang Cao}, \bibinfo{person}{Haotian Wang}, \bibinfo{person}{Guang Wang}, \bibinfo{person}{Tian He}, {and} \bibinfo{person}{Desheng Zhang}.} \bibinfo{year}{2023}\natexlab{}.
\newblock \showarticletitle{AutoBuild: Automatic Community Building Labeling for Last-mile Delivery}. In \bibinfo{booktitle}{\emph{Proceedings of the 32nd ACM International Conference on Information and Knowledge Management}}. \bibinfo{pages}{4623--4630}.
\newblock


\bibitem[Hu et~al\mbox{.}(2020)]%
        {hu2020stochastic}
\bibfield{author}{\bibinfo{person}{Jilin Hu}, \bibinfo{person}{Bin Yang}, \bibinfo{person}{Chenjuan Guo}, \bibinfo{person}{Christian~S Jensen}, {and} \bibinfo{person}{Hui Xiong}.} \bibinfo{year}{2020}\natexlab{}.
\newblock \showarticletitle{Stochastic origin-destination matrix forecasting using dual-stage graph convolutional, recurrent neural networks}. In \bibinfo{booktitle}{\emph{2020 IEEE 36th International conference on data engineering (ICDE)}}. IEEE, \bibinfo{pages}{1417--1428}.
\newblock


\bibitem[Jin et~al\mbox{.}(2022)]%
        {jin2022stgnn}
\bibfield{author}{\bibinfo{person}{Guangyin Jin}, \bibinfo{person}{Min Wang}, \bibinfo{person}{Jinlei Zhang}, \bibinfo{person}{Hengyu Sha}, {and} \bibinfo{person}{Jincai Huang}.} \bibinfo{year}{2022}\natexlab{}.
\newblock \showarticletitle{STGNN-TTE: Travel time estimation via spatial--temporal graph neural network}.
\newblock \bibinfo{journal}{\emph{Future Generation Computer Systems}}  \bibinfo{volume}{126} (\bibinfo{year}{2022}), \bibinfo{pages}{70--81}.
\newblock


\bibitem[Jin et~al\mbox{.}(2021)]%
        {jin2021hierarchical}
\bibfield{author}{\bibinfo{person}{Guangyin Jin}, \bibinfo{person}{Huan Yan}, \bibinfo{person}{Fuxian Li}, \bibinfo{person}{Yong Li}, {and} \bibinfo{person}{Jincai Huang}.} \bibinfo{year}{2021}\natexlab{}.
\newblock \showarticletitle{Hierarchical neural architecture search for travel time estimation}. In \bibinfo{booktitle}{\emph{Proceedings of the 29th International Conference on Advances in Geographic Information Systems}}. \bibinfo{pages}{91--94}.
\newblock


\bibitem[Li et~al\mbox{.}(2023)]%
        {li2023traffic}
\bibfield{author}{\bibinfo{person}{Haoran Li}, \bibinfo{person}{Zhiqiang Lv}, \bibinfo{person}{Jianbo Li}, \bibinfo{person}{Zhihao Xu}, \bibinfo{person}{Yue Wang}, \bibinfo{person}{Haokai Sun}, {and} \bibinfo{person}{Zhaoyu Sheng}.} \bibinfo{year}{2023}\natexlab{}.
\newblock \showarticletitle{Traffic Flow Forecasting in the COVID-19: A Deep Spatial-Temporal Model Based on Discrete Wavelet Transformation}.
\newblock \bibinfo{journal}{\emph{ACM Transactions on Knowledge Discovery from Data}} \bibinfo{volume}{17}, \bibinfo{number}{5} (\bibinfo{year}{2023}), \bibinfo{pages}{1--28}.
\newblock


\bibitem[Mao et~al\mbox{.}(2023)]%
        {mao2023drl4route}
\bibfield{author}{\bibinfo{person}{Xiaowei Mao}, \bibinfo{person}{Haomin Wen}, \bibinfo{person}{Hengrui Zhang}, \bibinfo{person}{Huaiyu Wan}, \bibinfo{person}{Lixia Wu}, \bibinfo{person}{Jianbin Zheng}, \bibinfo{person}{Haoyuan Hu}, {and} \bibinfo{person}{Youfang Lin}.} \bibinfo{year}{2023}\natexlab{}.
\newblock \showarticletitle{Drl4route: A deep reinforcement learning framework for pick-up and delivery route prediction}. In \bibinfo{booktitle}{\emph{Proceedings of the 29th ACM SIGKDD Conference on Knowledge Discovery and Data Mining}}. \bibinfo{pages}{4628--4637}.
\newblock


\bibitem[Ruan et~al\mbox{.}(2022)]%
        {ruan2022service}
\bibfield{author}{\bibinfo{person}{Sijie Ruan}, \bibinfo{person}{Cheng Long}, \bibinfo{person}{Zhipeng Ma}, \bibinfo{person}{Jie Bao}, \bibinfo{person}{Tianfu He}, \bibinfo{person}{Ruiyuan Li}, \bibinfo{person}{Yiheng Chen}, \bibinfo{person}{Shengnan Wu}, {and} \bibinfo{person}{Yu Zheng}.} \bibinfo{year}{2022}\natexlab{}.
\newblock \showarticletitle{Service Time Prediction for Delivery Tasks via Spatial Meta-Learning}. In \bibinfo{booktitle}{\emph{Proceedings of the 28th ACM SIGKDD Conference on Knowledge Discovery and Data Mining}}. \bibinfo{pages}{3829--3837}.
\newblock


\bibitem[Ruan et~al\mbox{.}(2020)]%
        {ruan2020doing}
\bibfield{author}{\bibinfo{person}{Sijie Ruan}, \bibinfo{person}{Zi Xiong}, \bibinfo{person}{Cheng Long}, \bibinfo{person}{Yiheng Chen}, \bibinfo{person}{Jie Bao}, \bibinfo{person}{Tianfu He}, \bibinfo{person}{Ruiyuan Li}, \bibinfo{person}{Shengnan Wu}, \bibinfo{person}{Zhongyuan Jiang}, {and} \bibinfo{person}{Yu Zheng}.} \bibinfo{year}{2020}\natexlab{}.
\newblock \showarticletitle{Doing in one go: delivery time inference based on couriers' trajectories}. In \bibinfo{booktitle}{\emph{Proceedings of the 26th ACM SIGKDD International Conference on Knowledge Discovery \& Data Mining}}. \bibinfo{pages}{2813--2821}.
\newblock


\bibitem[Sui et~al\mbox{.}(2024)]%
        {sui2024congestion}
\bibfield{author}{\bibinfo{person}{Hongjie Sui}, \bibinfo{person}{Huan Yan}, \bibinfo{person}{Tianyi Zheng}, \bibinfo{person}{Wenzhen Huang}, \bibinfo{person}{Yunlin Zhuang}, {and} \bibinfo{person}{Yong Li}.} \bibinfo{year}{2024}\natexlab{}.
\newblock \showarticletitle{Congestion-aware Spatio-Temporal Graph Convolutional Network Based A* Search Algorithm for Fastest Route Search}.
\newblock \bibinfo{journal}{\emph{ACM Transactions on Knowledge Discovery from Data}} (\bibinfo{year}{2024}).
\newblock


\bibitem[Tang et~al\mbox{.}(2020)]%
        {tang2020joint}
\bibfield{author}{\bibinfo{person}{Xianfeng Tang}, \bibinfo{person}{Huaxiu Yao}, \bibinfo{person}{Yiwei Sun}, \bibinfo{person}{Charu Aggarwal}, \bibinfo{person}{Prasenjit Mitra}, {and} \bibinfo{person}{Suhang Wang}.} \bibinfo{year}{2020}\natexlab{}.
\newblock \showarticletitle{Joint modeling of local and global temporal dynamics for multivariate time series forecasting with missing values}. In \bibinfo{booktitle}{\emph{Proceedings of the AAAI Conference on Artificial Intelligence}}, Vol.~\bibinfo{volume}{34}. \bibinfo{pages}{5956--5963}.
\newblock


\bibitem[Tianqi(2023)]%
        {lishitianqi}
\bibfield{author}{\bibinfo{person}{Tianqi}.} \bibinfo{year}{2023}\natexlab{}.
\newblock \bibinfo{title}{Weather Forecast}.
\newblock
\newblock
\urldef\tempurl%
\url{http://lishi.tianqi.com/}
\showURL{%
\tempurl}


\bibitem[Vaswani et~al\mbox{.}(2017)]%
        {vaswani2017attention}
\bibfield{author}{\bibinfo{person}{Ashish Vaswani}, \bibinfo{person}{Noam Shazeer}, \bibinfo{person}{Niki Parmar}, \bibinfo{person}{Jakob Uszkoreit}, \bibinfo{person}{Llion Jones}, \bibinfo{person}{Aidan~N Gomez}, \bibinfo{person}{{\L}ukasz Kaiser}, {and} \bibinfo{person}{Illia Polosukhin}.} \bibinfo{year}{2017}\natexlab{}.
\newblock \showarticletitle{Attention is all you need}.
\newblock \bibinfo{journal}{\emph{Advances in neural information processing systems}}  \bibinfo{volume}{30} (\bibinfo{year}{2017}).
\newblock


\bibitem[Vinyals et~al\mbox{.}(2015)]%
        {vinyals2015pointer}
\bibfield{author}{\bibinfo{person}{Oriol Vinyals}, \bibinfo{person}{Meire Fortunato}, {and} \bibinfo{person}{Navdeep Jaitly}.} \bibinfo{year}{2015}\natexlab{}.
\newblock \showarticletitle{Pointer networks}.
\newblock \bibinfo{journal}{\emph{Advances in neural information processing systems}}  \bibinfo{volume}{28} (\bibinfo{year}{2015}).
\newblock


\bibitem[Wang et~al\mbox{.}(2019)]%
        {wang2019simple}
\bibfield{author}{\bibinfo{person}{Hongjian Wang}, \bibinfo{person}{Xianfeng Tang}, \bibinfo{person}{Yu-Hsuan Kuo}, \bibinfo{person}{Daniel Kifer}, {and} \bibinfo{person}{Zhenhui Li}.} \bibinfo{year}{2019}\natexlab{}.
\newblock \showarticletitle{A simple baseline for travel time estimation using large-scale trip data}.
\newblock \bibinfo{journal}{\emph{ACM Transactions on Intelligent Systems and Technology (TIST)}} \bibinfo{volume}{10}, \bibinfo{number}{2} (\bibinfo{year}{2019}), \bibinfo{pages}{1--22}.
\newblock


\bibitem[Wang et~al\mbox{.}(2018)]%
        {wang2018learning}
\bibfield{author}{\bibinfo{person}{Zheng Wang}, \bibinfo{person}{Kun Fu}, {and} \bibinfo{person}{Jieping Ye}.} \bibinfo{year}{2018}\natexlab{}.
\newblock \showarticletitle{Learning to estimate the travel time}. In \bibinfo{booktitle}{\emph{Proceedings of the 24th ACM SIGKDD International Conference on Knowledge Discovery \& Data Mining}}. \bibinfo{pages}{858--866}.
\newblock


\bibitem[Wang et~al\mbox{.}(2022)]%
        {wang2022event}
\bibfield{author}{\bibinfo{person}{Zhaonan Wang}, \bibinfo{person}{Renhe Jiang}, \bibinfo{person}{Hao Xue}, \bibinfo{person}{Flora~D Salim}, \bibinfo{person}{Xuan Song}, {and} \bibinfo{person}{Ryosuke Shibasaki}.} \bibinfo{year}{2022}\natexlab{}.
\newblock \showarticletitle{Event-aware multimodal mobility nowcasting}. In \bibinfo{booktitle}{\emph{Proceedings of the AAAI Conference on Artificial Intelligence}}, Vol.~\bibinfo{volume}{36}. \bibinfo{pages}{4228--4236}.
\newblock


\bibitem[Wen et~al\mbox{.}(2023a)]%
        {wen2023modeling}
\bibfield{author}{\bibinfo{person}{Haomin Wen}, \bibinfo{person}{Youfang Lin}, \bibinfo{person}{Yuxuan Hu}, \bibinfo{person}{Fan Wu}, \bibinfo{person}{Mingxuan Xia}, \bibinfo{person}{Xinyi Zhang}, \bibinfo{person}{Lixia Wu}, \bibinfo{person}{Haoyuan Hu}, {and} \bibinfo{person}{Huaiyu Wan}.} \bibinfo{year}{2023}\natexlab{a}.
\newblock \showarticletitle{Modeling Spatial--Temporal Constraints and Spatial-Transfer Patterns for Couriers’ Package Pick-up Route Prediction}.
\newblock \bibinfo{journal}{\emph{IEEE Transactions on Intelligent Transportation Systems}} (\bibinfo{year}{2023}).
\newblock


\bibitem[Wen et~al\mbox{.}(2022a)]%
        {wen2022graph2route}
\bibfield{author}{\bibinfo{person}{Haomin Wen}, \bibinfo{person}{Youfang Lin}, \bibinfo{person}{Xiaowei Mao}, \bibinfo{person}{Fan Wu}, \bibinfo{person}{Yiji Zhao}, \bibinfo{person}{Haochen Wang}, \bibinfo{person}{Jianbin Zheng}, \bibinfo{person}{Lixia Wu}, \bibinfo{person}{Haoyuan Hu}, {and} \bibinfo{person}{Huaiyu Wan}.} \bibinfo{year}{2022}\natexlab{a}.
\newblock \showarticletitle{Graph2route: A dynamic spatial-temporal graph neural network for pick-up and delivery route prediction}. In \bibinfo{booktitle}{\emph{Proceedings of the 28th ACM SIGKDD Conference on Knowledge Discovery and Data Mining}}. \bibinfo{pages}{4143--4152}.
\newblock


\bibitem[Wen et~al\mbox{.}(2022b)]%
        {wen2022deeproute+}
\bibfield{author}{\bibinfo{person}{Haomin Wen}, \bibinfo{person}{Youfang Lin}, \bibinfo{person}{Huaiyu Wan}, \bibinfo{person}{Shengnan Guo}, \bibinfo{person}{Fan Wu}, \bibinfo{person}{Lixia Wu}, \bibinfo{person}{Chao Song}, {and} \bibinfo{person}{Yinghui Xu}.} \bibinfo{year}{2022}\natexlab{b}.
\newblock \showarticletitle{DeepRoute+: Modeling Couriers’ Spatial-temporal Behaviors and Decision Preferences for Package Pick-up Route Prediction}.
\newblock \bibinfo{journal}{\emph{ACM Transactions on Intelligent Systems and Technology (TIST)}} \bibinfo{volume}{13}, \bibinfo{number}{2} (\bibinfo{year}{2022}), \bibinfo{pages}{1--23}.
\newblock


\bibitem[Wen et~al\mbox{.}(2021)]%
        {wen2021package}
\bibfield{author}{\bibinfo{person}{Haomin Wen}, \bibinfo{person}{Youfang Lin}, \bibinfo{person}{Fan Wu}, \bibinfo{person}{Huaiyu Wan}, \bibinfo{person}{Shengnan Guo}, \bibinfo{person}{Lixia Wu}, \bibinfo{person}{Chao Song}, {and} \bibinfo{person}{Yinghui Xu}.} \bibinfo{year}{2021}\natexlab{}.
\newblock \showarticletitle{Package pick-up route prediction via modeling couriers’ spatial-temporal behaviors}. In \bibinfo{booktitle}{\emph{2021 IEEE 37th International Conference on Data Engineering (ICDE)}}. IEEE, \bibinfo{pages}{2141--2146}.
\newblock


\bibitem[Wen et~al\mbox{.}(2023c)]%
        {wen2023enough}
\bibfield{author}{\bibinfo{person}{Haomin Wen}, \bibinfo{person}{Youfang Lin}, \bibinfo{person}{Fan Wu}, \bibinfo{person}{Huaiyu Wan}, \bibinfo{person}{Zhongxiang Sun}, \bibinfo{person}{Tianyue Cai}, \bibinfo{person}{Hongyu Liu}, \bibinfo{person}{Shengnan Guo}, \bibinfo{person}{Jianbin Zheng}, \bibinfo{person}{Chao Song}, {et~al\mbox{.}}} \bibinfo{year}{2023}\natexlab{c}.
\newblock \showarticletitle{Enough Waiting for the Couriers: Learning to Estimate Package Pick-up Arrival Time from Couriers’ Spatial-Temporal Behaviors}.
\newblock \bibinfo{journal}{\emph{ACM Transactions on Intelligent Systems and Technology}} \bibinfo{volume}{14}, \bibinfo{number}{3} (\bibinfo{year}{2023}), \bibinfo{pages}{1--22}.
\newblock


\bibitem[Wen et~al\mbox{.}(2023b)]%
        {wen2023survey}
\bibfield{author}{\bibinfo{person}{Haomin Wen}, \bibinfo{person}{Youfang Lin}, \bibinfo{person}{Lixia Wu}, \bibinfo{person}{Xiaowei Mao}, \bibinfo{person}{Tianyue Cai}, \bibinfo{person}{Yunfeng Hou}, \bibinfo{person}{Shengnan Guo}, \bibinfo{person}{Yuxuan Liang}, \bibinfo{person}{Guangyin Jin}, \bibinfo{person}{Yiji Zhao}, {et~al\mbox{.}}} \bibinfo{year}{2023}\natexlab{b}.
\newblock \showarticletitle{A Survey on Service Route and Time Prediction in Instant Delivery: Taxonomy, Progress, and Prospects}.
\newblock \bibinfo{journal}{\emph{arXiv preprint arXiv:2309.01194}} (\bibinfo{year}{2023}).
\newblock


\bibitem[Wikipedia(2023)]%
        {geocoding}
\bibfield{author}{\bibinfo{person}{Wikipedia}.} \bibinfo{year}{2023}\natexlab{}.
\newblock \bibinfo{title}{Address geocoding}.
\newblock
\newblock
\urldef\tempurl%
\url{https://en.wikipedia.org/wiki/Address\_geocoding}
\showURL{%
\tempurl}


\bibitem[Wu and Wu(2019)]%
        {wu2019deepeta}
\bibfield{author}{\bibinfo{person}{Fan Wu} {and} \bibinfo{person}{Lixia Wu}.} \bibinfo{year}{2019}\natexlab{}.
\newblock \showarticletitle{Deepeta: A spatial-temporal sequential neural network model for estimating time of arrival in package delivery system}. In \bibinfo{booktitle}{\emph{Proceedings of the AAAI conference on artificial intelligence}}, Vol.~\bibinfo{volume}{33}. \bibinfo{pages}{774--781}.
\newblock


\bibitem[Yan et~al\mbox{.}(2022)]%
        {yan2022jointly}
\bibfield{author}{\bibinfo{person}{Huan Yan}, \bibinfo{person}{Guangyin Jin}, \bibinfo{person}{Deng Wang}, \bibinfo{person}{Yue Liu}, {and} \bibinfo{person}{Yong Li}.} \bibinfo{year}{2022}\natexlab{}.
\newblock \showarticletitle{Jointly Modeling Intersections and Road Segments for Travel Time Estimation via Dual Graph Convolutional Networks}. In \bibinfo{booktitle}{\emph{International Conference on Spatial Data and Intelligence}}. Springer, \bibinfo{pages}{19--34}.
\newblock


\bibitem[Yi et~al\mbox{.}(2023)]%
        {yi2023deepsta}
\bibfield{author}{\bibinfo{person}{Jinhui Yi}, \bibinfo{person}{Huan Yan}, \bibinfo{person}{Haotian Wang}, \bibinfo{person}{Jian Yuan}, {and} \bibinfo{person}{Yong Li}.} \bibinfo{year}{2023}\natexlab{}.
\newblock \showarticletitle{DeepSTA: A Spatial-Temporal Attention Network for Logistics Delivery Timely Rate Prediction in Anomaly Conditions}. In \bibinfo{booktitle}{\emph{Proceedings of the 32nd ACM International Conference on Information and Knowledge Management}}. \bibinfo{pages}{4916--4922}.
\newblock


\bibitem[Yi et~al\mbox{.}(2024)]%
        {yi2024rccnet}
\bibfield{author}{\bibinfo{person}{Jinhui Yi}, \bibinfo{person}{Huan Yan}, \bibinfo{person}{Haotian Wang}, \bibinfo{person}{Jian Yuan}, {and} \bibinfo{person}{Yong Li}.} \bibinfo{year}{2024}\natexlab{}.
\newblock \showarticletitle{RCCNet: A Spatial-Temporal Neural Network Model for Logistics Delivery Timely Rate Prediction}.
\newblock \bibinfo{journal}{\emph{ACM Transactions on Intelligent Systems and Technology}} (\bibinfo{year}{2024}).
\newblock


\bibitem[Yuan et~al\mbox{.}(2020)]%
        {yuan2020effective}
\bibfield{author}{\bibinfo{person}{Haitao Yuan}, \bibinfo{person}{Guoliang Li}, \bibinfo{person}{Zhifeng Bao}, {and} \bibinfo{person}{Ling Feng}.} \bibinfo{year}{2020}\natexlab{}.
\newblock \showarticletitle{Effective travel time estimation: When historical trajectories over road networks matter}. In \bibinfo{booktitle}{\emph{Proceedings of the 2020 acm sigmod international conference on management of data}}. \bibinfo{pages}{2135--2149}.
\newblock


\bibitem[Zhang et~al\mbox{.}(2023b)]%
        {zhang2023efficient}
\bibfield{author}{\bibinfo{person}{Guozhen Zhang}, \bibinfo{person}{Jinhui Yi}, \bibinfo{person}{Jian Yuan}, \bibinfo{person}{Yong Li}, {and} \bibinfo{person}{Depeng Jin}.} \bibinfo{year}{2023}\natexlab{b}.
\newblock \showarticletitle{Das: Efficient street view image sampling for urban prediction}.
\newblock \bibinfo{journal}{\emph{ACM Transactions on Intelligent Systems and Technology}} \bibinfo{volume}{14}, \bibinfo{number}{2} (\bibinfo{year}{2023}), \bibinfo{pages}{1--20}.
\newblock


\bibitem[Zhang et~al\mbox{.}(2023a)]%
        {zhang2023dual}
\bibfield{author}{\bibinfo{person}{Lei Zhang}, \bibinfo{person}{Mingliang Wang}, \bibinfo{person}{Xin Zhou}, \bibinfo{person}{Xingyu Wu}, \bibinfo{person}{Yiming Cao}, \bibinfo{person}{Yonghui Xu}, \bibinfo{person}{Lizhen Cui}, {and} \bibinfo{person}{Zhiqi Shen}.} \bibinfo{year}{2023}\natexlab{a}.
\newblock \showarticletitle{Dual Graph Multitask Framework for Imbalanced Delivery Time Estimation}. In \bibinfo{booktitle}{\emph{International Conference on Database Systems for Advanced Applications}}. Springer, \bibinfo{pages}{606--618}.
\newblock


\bibitem[Zhang et~al\mbox{.}(2023c)]%
        {zhang2023delivery}
\bibfield{author}{\bibinfo{person}{Lei Zhang}, \bibinfo{person}{Xin Zhou}, \bibinfo{person}{Zhiwei Zeng}, \bibinfo{person}{Yiming Cao}, \bibinfo{person}{Yonghui Xu}, \bibinfo{person}{Mingliang Wang}, \bibinfo{person}{Xingyu Wu}, \bibinfo{person}{Yong Liu}, \bibinfo{person}{Lizhen Cui}, {and} \bibinfo{person}{Zhiqi Shen}.} \bibinfo{year}{2023}\natexlab{c}.
\newblock \showarticletitle{Delivery time prediction using large-scale graph structure learning based on quantile regression}. In \bibinfo{booktitle}{\emph{2023 IEEE 39th International Conference on Data Engineering (ICDE)}}. IEEE, \bibinfo{pages}{3403--3416}.
\newblock


\bibitem[Zhang et~al\mbox{.}(2020)]%
        {zhang2020real}
\bibfield{author}{\bibinfo{person}{Wen Zhang}, \bibinfo{person}{Yang Wang}, \bibinfo{person}{Xike Xie}, \bibinfo{person}{Chuancai Ge}, {and} \bibinfo{person}{Hengchang Liu}.} \bibinfo{year}{2020}\natexlab{}.
\newblock \showarticletitle{Real-time travel time estimation with sparse reliable surveillance information}.
\newblock \bibinfo{journal}{\emph{Proceedings of the ACM on Interactive, Mobile, Wearable and Ubiquitous Technologies}} \bibinfo{volume}{4}, \bibinfo{number}{1} (\bibinfo{year}{2020}), \bibinfo{pages}{1--23}.
\newblock


\bibitem[Zhou et~al\mbox{.}(2023)]%
        {zhou2023inductive}
\bibfield{author}{\bibinfo{person}{Xin Zhou}, \bibinfo{person}{Jinglong Wang}, \bibinfo{person}{Yong Liu}, \bibinfo{person}{Xingyu Wu}, \bibinfo{person}{Zhiqi Shen}, {and} \bibinfo{person}{Cyril Leung}.} \bibinfo{year}{2023}\natexlab{}.
\newblock \showarticletitle{Inductive graph transformer for delivery time estimation}. In \bibinfo{booktitle}{\emph{Proceedings of the Sixteenth ACM International Conference on Web Search and Data Mining}}. \bibinfo{pages}{679--687}.
\newblock


\bibitem[Zhu et~al\mbox{.}(2020)]%
        {zhu2020order}
\bibfield{author}{\bibinfo{person}{Lin Zhu}, \bibinfo{person}{Wei Yu}, \bibinfo{person}{Kairong Zhou}, \bibinfo{person}{Xing Wang}, \bibinfo{person}{Wenxing Feng}, \bibinfo{person}{Pengyu Wang}, \bibinfo{person}{Ning Chen}, {and} \bibinfo{person}{Pei Lee}.} \bibinfo{year}{2020}\natexlab{}.
\newblock \showarticletitle{Order fulfillment cycle time estimation for on-demand food delivery}. In \bibinfo{booktitle}{\emph{Proceedings of the 26th ACM SIGKDD International Conference on Knowledge Discovery \& Data Mining}}. \bibinfo{pages}{2571--2580}.
\newblock


\end{thebibliography}

% \newpage
% \nobalance
\appendix

\section{IMPLEMENTATION DETAILs}

\subsection{Newly Dispatched Packages}\label{Newly Dispatched}
\begin{figure}[htbp]
\centering
\includegraphics[width=\linewidth]{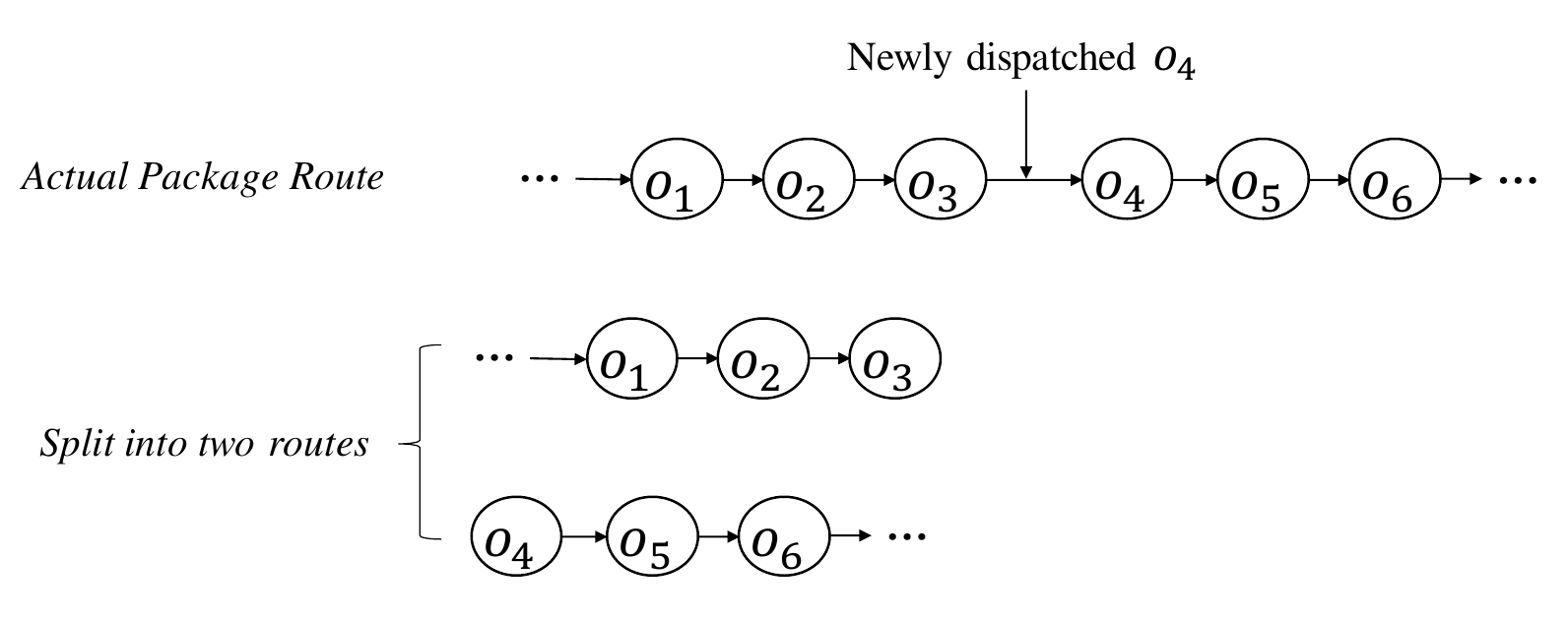}
\vspace{-5mm}
\caption{Split the route at the newly dispatched package.
} 
\label{split}
\vspace{-6mm}
\end{figure}

In the real world, delivery packages are pre-allocated to the courier before he/she commences his/her daily work, while pickup packages are assigned continually throughout the day. 
Whenever a pickup package is dispatched, the courier needs to re-plan the route, which will affect the pending packages consequently. 
Therefore, it is necessary to consider this issue when constructing the dataset.
As shown in Figure~\ref{split}, suppose that the courier has been dispatched delivery of $\{o_1,o_2,o_3,o_5,o_6\}$ at the beginning. After delivering $\{o_1,o_2,o_3\}$, the pickup of $o_4$ is dispatched to the courier. Thus he/she turns to pick up $o_4$ first, which results in the change in the sequence of order and time of $\{o_5,o_6\}$. 
To mitigate the impact of such unexpected dispatched packages, we split the original package route into two at $o_4$ when constructing datasets. %This ensures better management and allows for smoother handling of sudden pickup assignments.

\vspace{-3mm}

\end{document}